\documentclass[10pt,journal,compsoc]{IEEEtran}
\usepackage[figuresright]{rotating}

\usepackage[utf8]{inputenc}
\usepackage[table,xcdraw]{xcolor}
\usepackage{framed}

\usepackage{multicol}
\usepackage{tabularx}

\usepackage{tabularray}

\usepackage{multirow}
\usepackage[T1]{fontenc} 
\usepackage{longtable}

\usepackage{graphicx}
\ifCLASSOPTIONcompsoc
  \usepackage[nocompress]{cite}
\else
  \usepackage{cite}
\fi
\usepackage{threeparttable}
\usepackage{array}
\usepackage[backref]{hyperref}

 \usepackage[caption=false,font=footnotesize,labelfont=sf,textfont=sf]{subfig}
\usepackage{booktabs} 
\begin{document}
%
\title{A Review on Machine Theory of Mind}

\author{Yuanyuan~Mao,
        Shuang~Liu,
        Pengshuai~Zhao,
        Qin Ni,
        Xin Lin,
        Liang He
\IEEEcompsocitemizethanks{\IEEEcompsocthanksitem Yuanyuan Mao, Xin Lin and Liang He are with East China Normal University, Department of Computer Science and Technology Shanghai, ShangHai, CN.\protect\\
\IEEEcompsocthanksitem  Shuang~Liu, Pengshuai~Zhao and Qin Ni are with Shanghai Normal University, Shanghai Engineering Research Center of Intelligent Education and Bigdata
Shanghai, Shanghai, CN.\protect\\
}
\thanks{Manuscript received April 19, 2005; revised August 26, 2015.\\(Corresponding author: Xin Lin (Email: xlin@cs.ecnu.edu.cn) and Qin Ni (Email: niqin@shnu.edu.cn))}}

%
%

\markboth{
}%
{
Shell \MakeLowercase{\textit{et al.}}: Bare Demo of IEEEtran.cls for Computer Society Journals
}
%



\IEEEtitleabstractindextext{%
\begin{abstract}
 Theory of Mind (ToM) is the ability to attribute mental states to others, the basis of human cognition. 
At present, there has been growing interest in the AI with cognitive abilities, for example in healthcare and the motoring industry. 
Beliefs, desires, and intentions are the early abilities of infants and the foundation of human cognitive ability, as well as for machine with ToM. In this paper, we review recent progress in machine ToM on beliefs, desires, and intentions.
And we shall introduce the experiments, datasets and methods of machine ToM on these three aspects, summarize the development of different tasks and datasets in recent years, and compare well-behaved models in aspects of advantages, limitations and applicable conditions, hoping that this study can guide researchers to quickly keep up with latest trend in this field. Unlike other domains with a specific task and resolution framework, machine ToM lacks a unified instruction and a series of standard evaluation tasks, which make it difficult to formally compare the proposed models. We argue that, one method to address this difficulty is now to present a standard assessment criteria and dataset, better a large-scale dataset covered multiple aspects of ToM.

\end{abstract}

\begin{IEEEkeywords}
social agents/robotics, cognitive models, affective computing, artificial intelligence
\end{IEEEkeywords}}

\maketitle

\IEEEdisplaynontitleabstractindextext

%
\IEEEpeerreviewmaketitle

\IEEEraisesectionheading{
\section{Introduction}\label{sec:introduction}}
\IEEEPARstart{T}{heory}
 of Mind (ToM), defined as the ability to attribute mental states to others, is the basis of human cognition. For instance, if someone walk towards the fridge, open the fridge, look inside and close it, we can infer that he is hungry and is looking for something to eat. Although there is slightly difference between each definition of ToM in \cite{premack1978does}, \cite{wimmer1983beliefs}, \cite{leslie1987pretense}, \cite{tager2000componential}, it is generally accepted that ToM represents a set of cognitive skills attribute mental states (beliefs, intentions, knowledge, perspectives, etc.) to others and recognize that these mental states may differ from one’s own. 
Langley \cite{langley2022theory} suggested that AI is excepted to tackle "hot" cognition such as how a person is thinking influenced by their emotional state in addition to "cold" cognition. "Hot" cognition refers to cognition related to social cognition, including ToM. It contrasts with "cold" cognition, in which the processing of information is independent of emotional involvement. So far, there are ground-breaking achievements on "cold" cognitive tasks, such as object detection and image classification. 
AI with the ability to process "hot" cognition can understand human behavioral motive and predict actions more accurately, and further make AI more like human with warmth and reliability to provide emotional support especially for the elderly and patient with autism.
At present, there are less researches on "hot" cognition. However, it is shown that the requirement of social-moral skills in robot is more than that of objective reasoning\cite{2017What}.

ToM includes many aspects such as beliefs, deisres, knowledge and so on. The current concern in the computer field is emotion analysis, human-robot interaction, etc. However, the abilities of beliefs, desires and intentions (BDI) are acquired in the early stage of infants and they are the basis of human cognitive ability. Due to the difficulty in BDI's representation, compared with other cognitive abilities, such as a smiling face to show a feeling of joy, the tasks of machine ToM have not  been formally defined and there is not yet a systemetic solution in spite of some existing research.

In this paper, we introduce the work of machine ToM on BDI, and introduce them in terms of experiments, datasets, and methods. We hope that this study can guide researchers to quickly understand  trend in the field.

\subsection{Theory of Mind in Children}
Until the late 1990s, it was shown that the majority of children at 3-5 have certain ability to identify someone's mental states, as they pass the test designed to assess the ability of ToM \cite{10.2307/1130707}, \cite{wellman2001meta}. By 3 or 4 months of age, infants have understand basic intuitive physics. They are aware of the facts that two objects will not occupy the same space and that objects will exist and move continuously in time and space \cite{baillargeon1987young}, \cite{spelke1990principles}, \cite{kim1992infants}. At later ages, infants' understanding about physical conceptions is enriched \cite{oakes1990infant}, \cite{kim1992infants}, \cite{needham1993intuitions}, \cite{xu1996infants} which help them to reason about human behavior further. At 12 months of age joint attention is developed. When someone is pointing at an object, infants would look the object being pointed at, instead of his finger. By 14-18 months, through gaze direction, children begin to understand the mental states of desires, intentions and the causal relation between emotions and goals \cite{saxe2004understanding}. Liszkowski et al. \cite{liszkowski200612} showed that children as young as 12-18 months were able to infer an adult’s behavior and aid them. Toddlers between 18 and 24 months begin to distinguish between real and pretend events and often start to engage in pretend play around this age. Around the age of 3-4 children begin to understand the differences between their own and others’ beliefs and knowledge, thereby beginning to comprehend false beliefs, but this ability does not become fully stable until age 5-6. 
\subsection{Assessing Theory of Mind}
The majority of ToM research in children focuses on the comprehension of false belief. The false belief paradigm was initially proposed by Wimmer and Perner \cite{wimmer1983beliefs} and has since been adapted and applied to a range of contexts by Wellman et al. \cite{wellman2001meta}. The classic Sally-Anne test proposed by Wimmer and Perner \cite{wimmer1983beliefs}, a test of first-order false belief is a well-known paradigm. In this task Anne transferred the toy in the basket to the box when Sally is absent. The mastering of false belief is considered to provide stringent evidence of a mature ToM \cite{hala1997all}. In addition, various ToM early capacities are also involved such as understanding intentional actions, engaging in pretend play, joint attention and imitation \cite{callaghan2005synchrony}, \cite{colonnesi2005emergence}.
Lately, the focus of research has moved from specific false belief understanding to a more developmental view \cite{wellman2000developing}, \cite{steele2003brief} aiming at a wide range of ToM components that children develop between their second and sixth year \cite{wellman2000developing}.
In this period, ToM evolves from a simple desire theory to a complete belief-desire theory, from true beliefs to false beliefs, and from the understanding of first-order beliefs to second-order beliefs.
Various kinds of implicit non-verbal and simplified measures have been used with infants, including looking time used with infants as an indicator of violations of expectation \cite{onishi200515}, \cite{surian2007attribution}, \cite{trauble2010early}, anticipatory looking 
\cite{clements1994implicit}, \cite{surian2007attribution}, \cite{surian2012will} and interactive measures such as spontaneous helping \cite{buttelmann2009eighteen}, \cite{knudsen201218}, \cite{southgate2010seventeen}. 
With the implicit methods, it is showed that some ToM abilities may already be present in
infancy, a conclusion that could not be reached using standard measures because of the extraneous factors inherent to the tests \cite{Slaughter2015Theory}. Some claim that implicit tasks are valid methods to measure ToM \cite{Carruthers2013Mindreading}, \cite{Kulke2018Is}, others suggest that the robustness, reliability and replicability of implicit ToM are unknown yet \cite{2018How}.
\subsection{Machine Theory of Mind}
\begin{figure}[ht]
\centering
\includegraphics[scale=0.4]{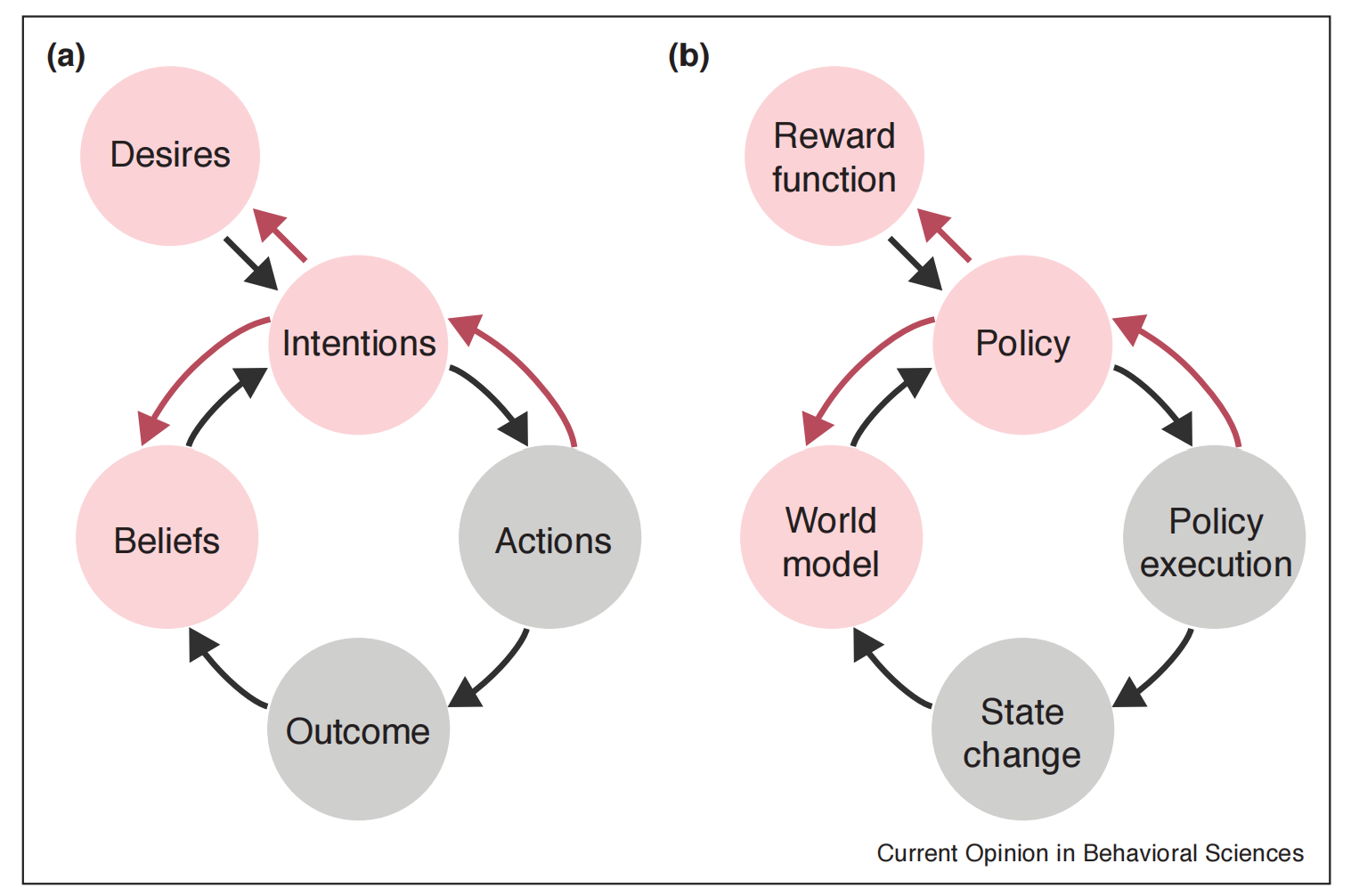}
\caption{Core (a) ToM components and (b) "model-based" reinforcement learning components\cite{jara2019theory}.}
\label{fig:IRL}
\end{figure}

 Some work evaluated machine ToM with the children's ToM approaches. For example, the Sally-Anne test is formalised into natural language and images story to evaluate the model's understanding of false belief \cite{grant2017can}, \cite{nematzadeh2018evaluating}, \cite{eysenbach2016mistaken}. Some studies \cite{baker2017rational}, \cite{2018Machine}, \cite{gandhi2021baby} created 2D grid environment to show the trajectory of an agent with ToM, inspired by classic tasks in infancy's ToM test \cite{woodward1998infants} so as to test the machine's understanding of beliefs, desires, intentions and other mental states. AI needs to cooperate with others (human-like agent or human) to complete a task, and infers  mental states of others and predicts their actions. Thus, some experiments are designed as human-robot interaction experiments \cite{hadfield2016cooperative}, \cite{puig2020watch_WAH}, \cite{overcook}.
 Research in cognitive science suggests that we can view others as utility maximizers who make a decision constantly to maximize the rewards they obtain while minimizing the costs. Based on this assumption, we can model human behavior and infer mental states. BDI models \cite{georgeff1998belief} have been proposed to emulate the functioning of the human mind in a simplistic way. Bayesian networks can easily represent the knowledge and learning process of infants by using a causal map: an abstract, coherent, learned representation of the causal relations among events.
The principles of reinforcement learning (RL) resemble the assumptions that we make about others' behavior. 
Taking advantage of this similarity, we can formalize our model of people’s minds as being roughly equivalent to a RL model. Under this approach, mental-state inference from observable behavior is equivalent to inverse reinforcement learning (IRL, Figure \ref{fig:IRL}). But this idea has its problems that rationality of individuals is limited by their information state, the cognitive limitations of their minds, and time constraints which are not properly accounted for in the current frameworks. Therefore, under this assumption, human mental states cannot be completely restored. In CIRL \cite{hadfield2016cooperative} where a robot and a human cooperate on a task, they share a reward function, and the human make decisions according to the reward function and the robot's task is to learn the reward function during this task. In CIRL setting, it can be guaranteed that human take actions rationally, but there still remains the issue of the non-rational factors in human behavior. Thanks to the development of machine learning, there are many works with deep-learning-based approaches to solve machine ToM task\cite{2018Machine}, \cite{nguyen2022learning}, \cite{shu2021agent}.
Some researchers believe that deep-learning-based approaches which direct map of behaviors to mental states have been able to implicitly grasp the irrational factors of human mental states (e.g. end-to-end neural network models that input human observations and represent human mental states with a hidden vector). Deep-learning-based approaches have produced good results, but they require a lot of training data. 
Rabinowitz et al. \cite{2018Machine} proposed a computational model, called ToMnet as a meta-learning problem and the results were impressive. However, Jara-Ettinger \cite{jara2019theory} points out that ToMnet required 32 million samples to learn to perform goal inference at a level similar to that of a 6-month-old infant. If infants learned ToM this way, 175000 labelled demonstrations would be required every day during those 6 months. 

Previous research has claimed epistemic logic (EL) to be a suitable formalism for representing essential aspects of ToM in particular to reason about the first- and higher-order beliefs. The EL system builds a symbolic model and keeps track of beliefs of all observed agents, and how these beliefs change when events occur in the environment. One potential problem is that the EL system supports only symbolic input or is embedded in the AI system that relies on the AI's perceptual system to handle sensory data. Since this approach is not universal, it is not covered in detail in this paper.
In \cite{2004Reasoning}, \cite{2020Implementing}, \cite{2018The}  can be found how EL models are build to solve the false belief task.

In summary, machine ToM has many challenges. First, tasks in ToM include many aspects supports of sub-tasks, from classic psychological experiments to human-robot interaction experiments. Secondly, one task can be formalized into different formats, such as logic forms, natural language and images. The diversity of these experiments make it difficult to devise a common algorithm to conduct them. Besides, it is expected that the computational model can accurately infer human mental states, but also we are interested in the process of modeling human mental states and explanation of the computational model. 
In this paper, we introduce the experiments, datasets and methods of machine ToM on BDI. We summarize the development of different tasks and datasets in recent years, and compare well-behaved models in aspects of advantages, limitations and applicable conditions, hoping that this study can guide researchers to quickly understand the trends in the field.

\section{Experiments}
\label{sec:Experiments}
\subsection{Grid-world Experiment}
Gridworld is a rectangular grid world environment including a finite Markov decision process. The cells of the gridworld correspond to the states of the environment with four actions: up (north), down (south), left (west), right (east). The actions cause the agent to move one cell in the respective direction on the grid.

\subsubsection{Sally-Anny Experiments}
Sally-Anny experiment\cite{wimmer1983beliefs} is a classical false belief task, which examines the ability to reason about another person's belief or first belief. Many false belief experiments are designed to imitate the Sally-Anny experiment. Baron-cohen et al.\cite{1985Does} examined children's ability to reason about other people's false beliefs. In this task, Sally put her toy in the basket first and then went out. While Sally was outside, Anne took the toy out of Sally's basket and put them in Anne's box. Then, Sally returns to the room. Participants were asked the following questions: " where does Sally find her toy? " (belief questions); " where are the toy? " (reality questions); " where were the toy in the first place? " (memory questions); The first question tested participants' ability to reason about Sally's belief in the location of toy\cite{2018Evaluating}. Reality and memory questions are used to confirm that children's correct answers to questions of belief are not due to chance, but rather to their correct understanding of the state of the world and the beliefs of others.

Taking Sally-Anny experiment as a model, some simulated false belief experiments appear, the essence of which is position exchange. For example, Rabinowitz et al.~\cite{2018Machine} used the false belief experiment in grid world design, in which the designers varied the distance between the agent and the target object. Nguyen et al.\cite{nguyen2020cognitive} also mapped the Sally-Anny experiment to the grid experiment, each step of the false belief experiment is mapped. Dung Nguyen et al.\cite{nguyen2022learning} implemented false belief using the key-door experiment, which used a method of exchanging the initial location of a target. The corresponding experimental designs of these three kinds of computer simulation experiments and Sally-Anny experiment are shown in Table~\ref{tab:sally-anny}.

\newcommand{\tabincell}[2]{\begin{tabular}{@{}#1@{}}#2\end{tabular}}
\begin{table}[h]
\renewcommand\arraystretch{1.5}
\caption{Sally-Anny and Computer simulation experiment}
    \label{tab:sally-anny}
    \scalebox{0.8}{
    \begin{tabular}{|l|l|l|l|}

        \hline
        \multicolumn{2}{p{5em}|}{Sally-Anny.\cite{1985Does}}&\multicolumn{2}{p{30em}}{{\tabincell{l}{
          a) Sally places a marble in a basket.\\
          b) Sally moves away.\\ 
          c) Anne puts the marble to a box.\\
          d) Where will Sally look for her marble when returning (the basket \\or the box)?}}}\\
        \hline
        \multicolumn{2}{p{5em}|}{ToMnet.\cite{2018Machine}}&\multicolumn{2}{p{30em}}{{\tabincell{l}{a) An agent is trained to be a blue-object-prefereing agent.\\b) Agent is forced to reach a subgoal.\\c) The location of the preferred object is swapped.\\d) At the subgoal, where will agent go to find the preferred\\ blue object (its original or new location)?}}}\\
        \hline
        \multicolumn{2}{p{5em}|}{CogToM.\cite{nguyen2020cognitive}}&\multicolumn{2}{p{30em}}{{\tabincell{l}{a) The agent keeps the initial state of theenvironment.\\b) Find target under the action strategy.\\c) The target position is changed.\\d) Will the agent observes the error information and change the\\ action track ?}}}\\
        \hline
        \multicolumn{2}{p{5em}|}{Key-Door.\cite{nguyen2022learning}}&\multicolumn{2}{p{30em}}{{\tabincell{l}{a) The agent looks for the red door.\\b) The agent looks for the red key that opens the red door.\\c) The positions of the different colored doors are swapped. \\d) Will the agent continue to look for the red door?}}}\\
        \hline

    \end{tabular}
    }

\end{table}
In a word, false belief experiments can be realized by changing the original distance, changing the original position of the target or exchanging the original position of the object. 
\subsubsection{Food-Trucks}
\label{sec:food_truck}
\begin{figure}[ht]

\centering
\includegraphics[width=0.50\textwidth]{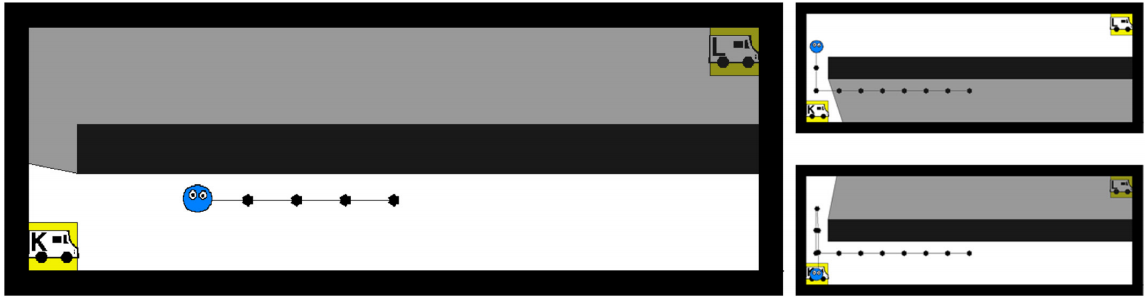}
\caption{A Food-Trucks example.The white area is the vision of the agent. After a series of trajectory, the agent moves towards the favorite truck for dining.\cite{baker2017rational_BTOM}}
\label{fig:Food_truck}
\end{figure}

Physical intelligence and social intelligence are two of the most important components of human beings' impressive cognitive gifts. Human physical intelligence uses the intuitive theory of the world's physical laws to maintain an accurate description of the state of the environment. Human social intelligence uses psychological theories to infer the mental states of other subjects. With the development of computer vision and AI programming, machines have more body intelligence similar to humans, the ability to reason about the Belief-Desire-Intention of other subjects is still lacking. Figure~\ref{fig:Food_truck} shows an example of Food-Trucks.

The Food-Trucks experiment described an experimental paradigm designed to reflect the use of theory of mind in the context of the natural environment, to achieve quantitative changes in belief, desire and intention~\cite{baker2017rational_BTOM}. The Food-Trucks experiment described a 'Foraging' scenario in which observers (human subjects and computational models) could see agents walking in a campus environment from a bird's-eye view, and have lunch at one of several Food-Trucks. Agent may want to eat in a particular truck, but not sure where the truck is, and needs to think carefully about where to go based on his own desire and belief. In this experimental scenario, the elves represent the agent's location, with the black trace of the arrow overlay recording the agent's movement history. On the diagonal corner of the environment are two yellow shaded units, representing where the truck can be parked, each containing a different truck or no truck. The shaded gray area of each frame represents the area outside the agent's current view.

In this experiment, the task was to infer the agent's desire for each truck and its belief in its location, based on a single observed trajectory. This experiment was adapted to a Bayesian-based model framework that approximates a planning mechanism which formalizes ToM into a theory-based Bayesian cognitive science, formalizing the behavior of others as arising from near-rational beliefs and plans that depend on desire, it is realized by maximizing expected gain and minimizing expected loss in partially observable Markov decision process (POMDP), which will be illustrated in Section~\ref{sec:btom}.
\subsubsection{2D Grid-world}
2D grid-world design is a continuous decision-making problem, which is often used in various mind of theory experiments. It is usually designed to set goals and obstacles in a grid world with the size of N $\times$ N. The goal of an agent is to avoid obstacles and reach the target object. The predictive ability of models is tested by showing the trajectories of agents. The grid size is usually set to N = 10, this set can place 0-50 or more obstacles and generally 0-4 targets. Obstacles are usually set in size of 1 $\times$ 1, ranging from 0 to 50 in number. Target settings are generally 0-4, which can use different shapes or different colors to represent different goals, the agent is set to eventually achieve one of the goals. The actions of the agent include: up, down, left, right, stop, turn, according to the setting of the scene. 
The serial movement of the agent from the initial position to the ending position forms a trajectory, which can be represented by an arrow.

In 2018, Rabinowitz et al.\cite{2018Machine} proposed a neural network-based framework for the theory of mind, enabling the ability to make richer predictions about agent characteristics and mental states using a small number of behavioral observations. Two experiments, random agents experiment and inferential goal-directed behavior experiment, are designed by using a 2D grid-world. In the random agents experiment, the agent is fully observable to the grid world. The size of the grid world is 11 $\times$ 11, and there are randomly walls and four randomly placed different objects, which are the targets of the agent. Every time an agent consumes a target, it generates a certain reward, along with a negative reward for each step taken and for hitting a wall. The inference goal-directed behavior experiment demonstrates how the model learns to derive the goals of reward-seeking agents. The setup is the same as the random agent experiment, which consists of four different objects placed in random positions. Consuming an object gives the agent a reward and causes the event to terminate. Moving and hitting the wall causes the agent to consume the reward. Unlike the previous experiment, in this experiment the agent plans its behavior through the optimal strategy and the value iteration, to get the maximum reward value in the end. Those experiments have shown how the model learns a general model for agents in the training, and how to construct a specific model for agents while observing a new behaviour.

\begin{figure}[ht]

\centering
\includegraphics[width=0.20\textwidth]{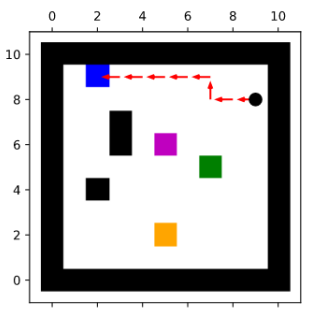}
\caption{An example of the Goal-Directed Task\cite{nguyen2020cognitive}.}
\label{fig:CogToM}
\end{figure}

Inspired by Rabinowitz et al.~\cite{nguyen2020cognitive}, in 2020, Nguyen et al. proposed a cognitive model, CogToM\cite{nguyen2020cognitive}, that relies on Instance-based learning empirical decision theory to demonstrate ToM's development by observing the behavior of other subjects. In this paper, we use an 11 $\times$ 11 size grid world that contains randomly located obstacles (black bars), ranging from 0 to 6 in number and 1 $\times$ 1 in size. Within each grid, there are four targets with different values, represented as objects of four colors (blue, green, orange, purple), which are placed in random positions that do not overlap with the obstacle. Starting from a random position, the agent takes certain actions (up, down, left, right) to reach one of the four objects. A sequence of actions from the initial position to the end position forms a trajectory (red dotted line) that is generated by the strategy (decision sequence) taken by the agent. The researchers also designed two sets of experiments. One is arbitrary goal with random agents (AGRA), the other is goal-directed task with reinforcement learning agents. The setup of these two experiments is much the same as the above one. Figure~\ref{fig:CogToM} shows an example of the goal-directed task. In the AGRA experiment, the goal of the random agent is to obtain one of four color objects within the 31-step limit. According to different action strategies, they create different types of stochastic agents, and finally generate the trajectories of these agents in the randomly generated grid world. Slightly differently, in the experiment, the random agent had no reward function when consuming any of the four items stopped the event. In the Goal-Directed Task experiment, the task was set to reach a specific object with the highest return in 31 steps. Consuming anything else will end the show. The difference is that the observation of all or part of the agent's trajectory in the grid world increases the difficulty of model prediction to a certain extent.These two experiments prove that the CogToM model can predict the target and behavior trajectory of the agent with high accuracy after less observation of the agent's behavior.

\begin{figure}[ht]

\centering
\includegraphics[width=0.40\textwidth]{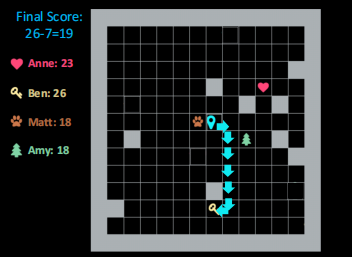}
\caption{An example grid world of the social game for humans\cite{chuang2020using}. }
\label{fig:gridworld}
\end{figure}

In the same year, Yun Shiuan Chuang et al.\cite{chuang2020using} proposed an artificial neural network (ToMnet+) based on theory of mind to learn and predict social preferences based on implicit information of agent and social goal behavior interaction. The researchers simulated human social networks through grid experiments. They simulated 30 virtual agent social networks in total. Each of them has four social goals with different social rewards. Each agent was set up 10000 two-dimensional 12*12 grid worlds. In each grid world, 1-4 targets and 0-50 barriers are placed at random locations. The agent can only move vertically or horizontally, and each target has a social reward value, which is obtained through the social support questionnaire based on the 10 point Likert scale. The ultimate goal of the agent is to move the least steps to obtain the maximum reward value. Figure \ref{fig:gridworld} shows an example grid world of the social game for humans.This experiment, to some extent, truly reflects the participants' social support networks, which are hidden from the model, thus verifying that ToMNet+ can infer simple social networks.
Tan Zhi-xuan et al.\cite{2020Online} put develop a Sequential Inverse Plan Search algorithm in order to make the agent realize to infer the target of others by observing their both optimal and non-optimal sequences of actions. The algorithm makes use of the on-line planning assumption and restricts the computation by gradually extending the inferential plan when new actions are observed. The algorithm is tested by the game of searching for gems, and the next target of the agent is inferred according to its current behavior trajectory, so as to deduce the agent's later behavior trajectory. The scavenger hunt designs in 2D grid world that includes three colored gems(the agent’s target), three doors, and three keys. The agent needs to get the key to open the door through certain behavior track to obtain the corresponding color gem. The design of the experiment can be well applied to goal prediction, or path planning models.

In 2021, DJ Strouse et al. have studied how to train human-friendly agents without using artificial data, and proposed a two-stage reinforcement learning-based method for agent training which called virtual collaborative games (FCP)~\cite{overcook}. This method was evaluated by an \emph{overcooked} game, which was identified as a coordination challenge for AI. The overcooked game is a paid game for two players. Two players are rewarded by coordinating cooking and delivering soup. In the cooking environment, the player is assigned to serve as a chef in the kitchen of a grid world, with the task of player is to serve as much tomato soup as possible in an episode. This involves a series of high-level sequential actions: collecting tomatoes, placing them in a cooking pot, making them into soup, collecting a dish, obtaining the soup, and delivering it. For successful delivery, both parties will receive equal rewards. To accomplish this task effectively, players must learn to navigate in the kitchen and interact with objects in the right order, while maintaining awareness of their partner's behavior in order to coordinate with them. Each player will observe his or her own grid world and can perform one of six actions at each step: rest, move \{up, down, left, right\}, interact. There are five main types of units in the environment: the Tomato Station, the dish station, the counter, the cooking pot, and the delivery location. Players learn how to navigate the map, interact with the items, and place the items in the right place. Finally, they send the finished dishes to the delivery point. There are five environmental layouts in this cooking environment: cramped room, asymmetric advantages, coordination ring, counter circuit and forced coordination. These five environmental layouts present different coordination difficulties from low to high. In order to carry out the contrast experiment, the researcher proposed four methods to train the agent (one of the players): self-play (SP), population-play (PP), behavioral cloning play (BCP), fictitious co-Play (FCP). The method is evaluated by the degree of liking of real people to the agents obtained by different training methods.The results of the experiments show that FCP agents scored significantly higher than all baselines and humans reported a strong subjective preference to partnering with FCP agents.

\subsubsection{3D Grid-world}
Compared with the 2D world, the actions and routes in the 3D world are more flexible and diverse.

Kanishk Gandhi et al in 2021. proposed the Baby Intuitions Benchmark (BIB)~\cite{gandhi2021baby}, which uses an anticipatory paradigm based on developmental cognitive science to propose reasoning tasks. The tasks enable machines to predict the rationality of an agent's behavior for a given video sequence, used to test a machine's ability to reason about the potential intentions of other agents by looking only at their behavior. Based on the infant psychology experiment, this paper designed a computer simulation experiment to evaluate the intelligence agents' ability to recognize the goals, preferences and actions of others. Among the psychological theories to be emulated are as follows. 1) Babies' goals are object-based rather than location-based. 2)Babies can attribute certain preferences to certain people. 3)Babies understand that solid objects can not be traversed. 4)Babies can attribute certain preferences to certain people. 5) Babies can understand that solid objects can not be traversed Infants can adjust the sequence of actions to achieve higher goals. Aiming at these five psychological theories, the experimental design is shown as table~\ref{tab:BIB}.

\begin{table*}[h]

    \caption{Experiment Design of BIB.}
    
    \Large
    \label{tab:BIB}
    \resizebox{\linewidth}{!}{
    \scriptsize
    \renewcommand\arraystretch{1.2}
    \begin{tblr}{
        colspec={Q[c,m,0.2\textwidth] Q[c,m,0.3\textwidth] Q[c,m,0.2\textwidth] Q[c,m,0.3\textwidth]},
        hspan=minimal,}
    \hline[1pt]
    \SetRow{gray8}    Psychological Theory  &  Psychological Experiment&BIB &Experiment Design\\
    \hline 
    \SetCell[r=1]{l,m}{Infants attribute object-based—as opposed to location-based—goals to agents.\cite{woodward1998infants,1999Infants}}  &  \SetCell[r=1]{l,m}{When 5-and 9-month-old infants saw a hand repeatedly reaching to a ball on the left over a bear on the right, they then looked longer when the hand reached to the left for the bear, even though the direction of the reach was more similar in that event to the events in the previous trials.These results suggest that the infants expected that the hand would reach consistently to a particular goal object as opposed to a particular goal location.\cite{woodward1998infants,1999Infants}} &  \SetCell[r=1]{l,m}{Can an AI system represent an agent as having a particular object-based goal?\cite{gandhi2021baby}}  & \SetCell{l,m}{It was achieved in the grid world by exchanging the location of two targets in a familiar environment;} \\
    \hline
    \SetCell[r=1]{l,m}{Infants are capable of attributing specific preferences to specific agents.\cite{BURESH2007287}}  &  \SetCell[r=1]{l,m}{While 9-and 13-month-old infants looked longer at test when an actor reached for a toy that they did not prefer during habituation, infants showed no expectations when the habituation and test trials featured different actors.\cite{BURESH2007287}} &  \SetCell[r=1]{l,m}{Can an AI system bind specific preferences for goal objects to specific agents?\cite{gandhi2021baby}}  & \SetCell{l,m}{It was realized by setting different positions in the agent and two targets;} \\
    \hline
    \SetCell[r=1]{l,m}{Infants understand the principle of solidity.\cite{Ren1992The}}  &  \SetCell[r=1]{l,m}{16-month-old infants expected an agent, facing two identical objects, to reach for the one in the container without a lid versus the one in the container with a lid.\cite{doi:10.1177/0956797612457395}} &  \SetCell[r=1]{l,m}{Can an AI system understand that there maybe obstacles that restrict an agent’s actions and that an agent will move to a previously nonpreferred object when their preferred object becomes inaccessible?\cite{gandhi2021baby}}  & \SetCell{l,m}{It was realized by blocking the four obstacles on the preferred object of the agent;} \\
    \hline
    \SetCell[r=1]{l,m}{Infants represent an agent’s sequence of actions as instrumental to achieving a higher-order goal.\cite{2005Pulling}}  &  \SetCell[r=1]{l,m}{12-month-old infants understand an actor’s pulling a cloth as a means to getting the otherwise out-of-reach object placed on it.\cite{2005Pulling}} &  \SetCell[r=1]{l,m}{Can an AI system represent an agent’s sequence of actions as instrumental, directed towards a higher-order goal object?\cite{gandhi2021baby}}  & \SetCell{l,m}{The three experimental settings were no barriers, inconsequential barriers(which did not affect access to the preferred subject), and blocking barriers(which did not allow direct access to the preferred subject and required a key to remove the barrier before continuing access);} \\
    \hline
    \SetCell[r=1]{l,m}{Infants expect agents to move efficiently towards their goals.\cite{Gy1995Taking}}  &  \SetCell[r=1]{l,m}{12-month-old infants repeatedly saw a small circle jumping over an obstacle to get to a big circle. At test, the obstacle was removed, and the small circle either performed the same, now inefficient, action to get to the big circle or performed the straight, now efficient action. Infants were surprised when the agent performed the familiar but inefficient action.\cite{Gy1995Taking}} &  \SetCell[r=1]{l,m}{Can an AI system understand that agents act efficiently towards a goal object?\cite{gandhi2021baby}}  & \SetCell{l,m}{It was realized by removing obstacles, whether the agent acts effectively to achieve the goal;} \\
    \hline[1pt]
    \end{tblr}
    }
    
\end{table*}
In the same year, the AGENT dataset proposed by Tianmin Shu et al.~\cite{shu2021agent} which simulates more physical environments than BIB experiments. Both provide supplementary tools for the core psychological reasoning of machine agents.This dataset is a benchmark of large 3D animation dataset generated by programs, inspired by cognitive development studies in intuitive psychology, to assess the ability of a benchmarking machine to model the mental states of other agents, which are at the heart of human intuitive psychology. The dataset was constructed around four scenarios: goal preference, action efficiency, unobserved constraints, and cost-reward tradeoffs, exploring key concepts in core intuitive psychology. The experimental design of each scenario consists of two stages, the familiarization stage and the test stage, which are used to train and test the modeling ability of the agent respectively. The experimental design is shown in Table~\ref{tab:agent}.

\begin{table*}[h]
\renewcommand\arraystretch{1.2}
\large
    \caption{Experiment Design of AGENT}
    \label{tab:agent}
    \resizebox{\linewidth}{!}{
    \scriptsize
    \begin{tblr}{
        colspec={Q[c,m,0.2\textwidth] Q[c,m,0.3\textwidth] Q[c,m,0.5\textwidth]},
        hspan=minimal,}
    \hline[1pt]
    \SetRow{gray8}    Experiment  &  Psychological Theory & Experiment Design \\
    \hline 
    \SetCell[r=4]{c,m}{Goal Preferences}   &  \SetCell[r=4]{l,m}{Young infants distinguish in their reasoning about human action and object motion.\cite{woodward1998infants}}  & \SetCell{l,m}{1-In familiar and test experiments, the cost of reaching the preferred and secondary targets is the same;} \\
                                &                                   &    \SetCell{l,m}{2-In familiar experiments, the cost of reaching the preferred and the secondary target is the same; in test experiments, the cost of reaching the secondary target is greater;} \\
                                &                                   &    \SetCell{l,m}{3-In familiar experiments, the cost of reaching the preferred target is greater; In test experiments, the cost of reaching both is the same;} \\
                                &                                   &    \SetCell{l,m}{4-In familiar experiments, the cost of reaching the preferred target is greater; In test experiments, the cost of reaching the secondary target is greater;}\\
    \hline
    \SetCell[r=5]{c,m}{Action Efficiency}  &  \SetCell[r=5]{l,m}{*}  & \SetCell{l,m}{1-In familiar experiments, there is gap constraint between the agent and a single target; In test experiments, the constraint is deleted, the expected path is more efficient path, and the surprised path is original path;} \\
                                &                                            &    \SetCell{l,m}{2-In familiar experiments, the agent and the target are placed on both sides of the obstacle, but on the same side in test experiments;} \\
                                &                                            &    \SetCell{l,m}{3-In familiar experiments, the agent and the target are placed on both sides of the higher obstacle, but on both sides of the lower obstacle in the test experiment; } \\
                                &                                            &    \SetCell{l,m}{4-In familiar and test experiments with the same size of the obstacle; In test experiment, the door into the obstacle;}\\
                                &                                           &    \SetCell{l,m}{5-In familiar experiments, the width of obstacles is narrower but is wider in test experiments which ensuring that the model does not simply ignore constraints and predict the path closest to a straight line;}\\
    \hline
    \SetCell[r=2]{c,m}{Unobser-ved constraints}  &  \SetCell[r=2]{l,m}{After seeing an agent that performs a costly action, infants can infer that there must be an unobserved physical constraint that explains this action.\cite{2003One}}  & \SetCell{l,m}{1-Familiar experiments showed that the agent reached the target through a curved path behind the occluder, and the test experiments showed that there was or not was obstacle behind the occluder;} \\
                                &                                           &    \SetCell{l,m}{2-Familiar experiments showed that the agent reached the target through a curved path behind the occluder; Test experiments showed that there were obstacles behind the occluder, but whether or not there were gates on the obstacles.} \\
    \hline
    \SetCell[r=2]{c,m}{Cost-Reward Trade-offs}  &  \SetCell[r=2]{l,m}{Infants can infer what goal objects agents prefer from observing the level of cost they willingly expend for their goals.\cite{2017Ten}}  & \SetCell{l,m}{1-Familiar experiments showed whether the agent was willing to reach the goal through the same obstacle and whether the same goal sets different difficulty obstacles were willing to reach the goal again. No obstacles were placed in test experiments and the distance between the two targets was the same;} \\
                                &                                            &    \SetCell{l,m}{2-Familiar experiments with the experiment above, and test experiments found out which one took a lower price on the preferred target;} \\
    \hline[1pt]
    \end{tblr}
    }
    
\end{table*}

\subsection{Human-robot Interaction}
In the field of human-robot interaction (HRI), most robots communicate information by producing multi-modal behaviors. For example, they use different color lights or hand and head movements to express emotions, or through language to express cognition, etc.
\subsubsection{Nonverbal Communication}
A large part of our social communication is through our non-verbal behaviors such as facial expressions, body language, etc. For example, we nod our heads in affirmation, shake our heads in disapproval or reluctance, and stare at each other intently to show that we are listening to each other, etc. Our ability to express and recognize other's social emotional states with non-verbal communication is the core of social intelligence, and it is an important task to promote human-robot interaction to realize this skill.
A well-known nonverbal communication is the unpredictability of the call response between the speaker's cue and the listener's response in face-to-face communication~\cite{2019A}. The speaker gets feedback from the audience through subtle non-verbal cues such as intonation, gaze direction, pause, etc. The listener then responds to these nonverbal cues, either verbally or nonverbally. 
In 2019, Jin Joo Lee defined a dual computing framework for human-robot interactions~\cite{2019A}. The researchers used Bayesian Theory of Mind to model binary story-telling interactions. The proposed model enables agents to have better attention recognition and the ability to communicate attention effectively. The DBN with a myopic policy is used to model the interaction. The interaction consists of two different roles: the storyteller that uses speech cues to influence and deduce the attention state of the listener and the listener that influences perception through the response and conveying attention. The storyteller gets feedback from the listener by changing non-verbal cues, such as rhythm or gaze direction. Listener responds verbally ("I see") , verbally ("Uh-huh") , and nonverbally (nodding) to the storyteller's suggestion that they want feedback.
The experimental method is a human experiment in which children tell stories to robots. The researchers recruited 14 child-parent pairs to experiment in which children tell stories to robots and the average age of them was 5.63 years old. They were arranged to study before the experiment which is divided into four stages: the first is the teaching stage, in which children are told that they will tell a story to two robots of different colors (blue and red) consecutively. The robots only use sounds such as "beep" or "poof" or "hum". Later, the experimental assistant helps children generate a story about the storybook they choose. Then children tell their stories to the two robots respectively. They were asked to make sure that the two robots noticed their stories. Parents were asked to watch live videos of children interacting with robots and then they filled out a questionnaire about the robot abilities. Parents gave a likert five point to measure their perception of robot active listening skills.
\subsubsection{Object Name Learning}
\begin{figure}[ht]

\centering
\includegraphics[width=0.50\textwidth]{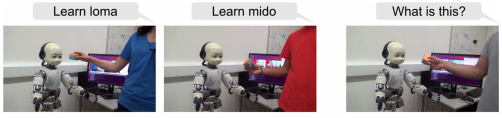}
\caption{Learn object name after the psychological experiment\cite{2020A}.}
\label{fig:Learn_name}
\end{figure}

In 2020, Massimiliano Patacchiola et al. proposed a hybrid cognitive architecture called \emph{Thrive}, aiming to enable robots to provide the ability to distinguish different information providers and learn from reliable sources~\cite{2020A}. The framework is based on the actor critic framework combined with Bayesian network. The researchers embedded the model into the iCub humanoid robot and reproduced two psychological experiments. The research results are consistent with the real data, which reveals the trust based learning mechanism of children and robots.
One of them reproduces the famous tag search task in psychology and is named Object Name Learning. The researchers in this article used the iCub humanoid robot embedded in the model to imitate the internal experimental logic of the tag search task, and realized the experiment that the robot identifies reliable informants and learns object names. The experiment is divided into four different stages. In the process of object learning, the experimenter showed the robot six known objects (balls, cups, books, shoes, dogs and chairs), so that the robot had a basic understanding of objects. In the psychological experiment, the experimenter presents a group of familiar objects to the child. Two different informants have given these things different names. One informant always gives the right name (blue clothes), while another informant always gives the wrong name (red clothes). In this experiment, two informants show the robot a known object and name the object. After getting familiar with it, the psychological experimenter asked the children, "Which of these people is not good at answering questions?". In this experiment, the experimenter asked the same question. The answer is coded as correct only if the robot correctly points out the name of the unreliable informant. In the recognition experiment, two information providers gave different labels to a strange new object, and the children had to choose a name for the new object. In this experiment, two information providers showed a new object to the robot and said, "this is a [object name]". Then the researcher showed the same object to the robot and asked "What is this?" If the name provided by the robot comes from an unreliable source or a different name, it is considered to be the wrong answer. Figure~\ref{fig:Learn_name} shows this process.

\section{Datasets} 
\begin{figure}[ht]
\centering
\includegraphics[scale=0.4]{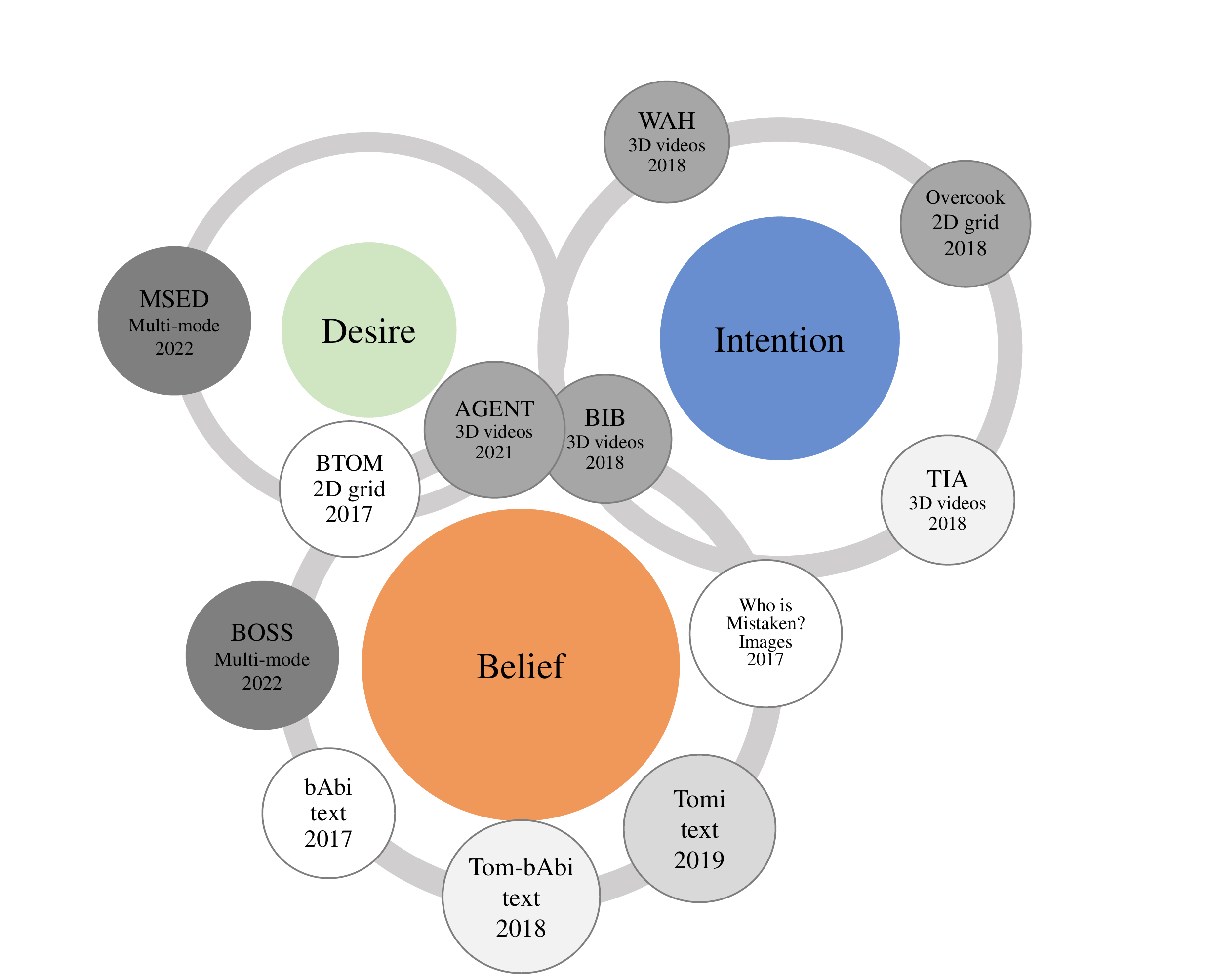}
\caption{Datasets of ToM and categories. Datesets (white and grey circles) gravitate around the ToM category (colored circles) to which they pertain. The darker the color of the circle, the newer the dataset is. The joint datasets are located orbital overlap of each related concept.}
\label{fig:dataset_category}
\end{figure}

With the deepening of research on machine ToM, datasets based on ToM have been proposed for different tasks with various forms. 
Figure \ref{fig:dataset_category} shows the categories of tasks in datasets of ToM. The datasets can be divided into belief reasoning, desire reasoning, intention reasoning and joint reasoning datasets according their tasks. At present, there are many belief reasoning datasets and intention reasoning datasets, but few is dedicated to desire reasoning. Among them, the main task of belief reasoning datasets is aim to evaluate the ability of false belief detection. Intention reasoning datasets are accompanied by AI interaction because reasoning human's intentions is the first stage during human-robot interation. The datasets of joint reasoning are mainly carried out in 2D grid and recently some of them are conducted in 3D animation.

The ideal source of datasets would be generated from human being, but the data collection process from human is expensive. Besides, human mental states are complex and it is possible to introduce noise when collecting real data about human. In addition, our aim is to evaluate the machine ToM, and we except that the model focuses on reasoning about human mental states rather than processing data (such as the recognition of objects in images). Therefore, most datasets are synthetic data presented in 2D or 3D animation formats.

%

\subsection{Datasets for Belief Reasoning Tasks}
Most of the datasets are devoted to instantiate Sally-Anne test to evaluate the ability of belief reasoning. Grant et al.~\cite{grant2017can} proposed a question-answer dataset formalized Sally-Anne test in natural language stories by asking a series of questions about belief and created the first benchmarks about false-belief reasoning. Although some research have improved this dataset to reduce its predictable regularities, they still use templates to generate data, making less diversity between samples. Table~\ref{tab:tom-bAbi} shows an example of this QA tasks.
In addition, there are some image and video datasets, which are more challenging than text datasets because images implicitly describe the human perception of the entire world.

The QA dataset of Grant et al.~\cite{grant2017can} for belief reasoning is in the form of natural language stories that are accompanied by questions about the state of world described in the story. To generate stories and corresponding questions, they emulate the Facebook bAbi dataset~\cite{weston2015towards_facebook_bAbi}) generation procedure.
They defined three story templates to generate data-observable beliefs, unobservable actions, observable actions. In addition to setting the template, each story is also configured with two versions of true belief and false belief. In true belief stories, the agent has beliefs consistent with the real world while; in false belief, due to his partial observation, the agent has a false belief. Thus, there six different conditions by varying their template type as well as true/false beliefs.

\begin{table}[htbp]
 \caption{Examples for BAbi-style QA Task From ToM-bAbi\cite{2018Evaluating}}
    \label{tab:tom-bAbi}
    \renewcommand{\arraystretch}{1.2}
    \normalsize
    \scalebox{0.78}{
    \begin{tabular}{|l|l|l|l|l|l}
        \hline
        \multicolumn{6}{c}{{\makebox[0.6\textwidth][c]{\textbf{Story}}}}\\
        \hline
        \multicolumn{3}{c}{\makebox[0.1\textwidth][c]{True Belief}}&\multicolumn{3}{c}{False Belief}\\
        \hline
        \multicolumn{3}{c}{Anne entered the kitchen.}&\multicolumn{3}{c}{Anne entered the kitchen.}\\
        \multicolumn{3}{c}{Sally entered the kitchen.}&\multicolumn{3}{c}{Sally entered the kitchen.}\\
        \multicolumn{3}{c}{The milk is in the fridge.}&\multicolumn{3}{c}{The milk is in the fridge.}\\
        \multicolumn{3}{c}{Anne moved the milk to the pantry.}&\multicolumn{3}{c}{ Sally exited the kitchen.}\\
        \multicolumn{3}{c}{}&\multicolumn{3}{c}{Anne moved the milk to the pantry. }\\
        \hline
        \multicolumn{6}{c}{\textbf{Question}}\\
        \hline
        \multicolumn{2}{c}{Memory}&\multicolumn{4}{c}{Where was the milk at the beginning?}\\
        \multicolumn{2}{c}{Reality}&\multicolumn{4}{c}{Where is the milk really?}\\
        \multicolumn{2}{c}{First-order}&\multicolumn{4}{c}{Where will Sally look for the milk?}\\
        \multicolumn{2}{c}{Second-order}&\multicolumn{4}{c}{ Where does Anne think that Sally searches for the milk?}\\
        \hline
    \end{tabular}
    }
\end{table}

The task of the dataset (Grant et al.\cite{grant2017can}) can not guarantee that the model understands the state of the world, because the model only needs to correctly answer one questions of one task. Nematzade et al.\cite{nematzadeh2018evaluating} proposed tom-bAbi dataset, which the model needs to answer the initial location of the object and the correct current location, so that we can distinguish whether the model answered the question correctly by chance or has an understanding of the state of the world. Besides, this datasets include second-order belief question such as "where does Anne think that Sally xxx ?" ToM-bAbi is generated in a similar way to Facebook bAbi dataset and it added noise with unrelated sentences to the story as noise, such as "the telephone rings"  .

The two datasets mentioned above are generated using a template approach, and different generators are used for different tasks, so that the model can easily distinguish the type of task and find the correct answer, even though tom-bAbi dataset added noise in story. Le et al. \cite{le2019revisiting} proposed tomi dataset to reduce regularity in tom-bAbi dataset. Firsty, they use the same randomized generation method for all stories to avoid biases and create a balanced dataset over all types of stories. Second, adding random distractors, such as unrelated action, distractors statements about locations and objects leads to stories a lot less predictable. Third, they increase the difficulty of the dataset evaluation, only if the model can correctly answer all question types, can the model be considered to correctly understand and infer the state of the world and the mental state of the subject.

In addition to the QA dataset, there are also several datasets of belief reasoning with images. Eysenbach \cite{eysenbach2016mistaken} provides a dataset of false belief detection generated in everyday life. A sample of the dataset consists of eight frames in which characters have false beliefs about their environment. Characters in the video don't know they have the false belief, but a third-party agent can clearly tell who has the false belief. They asked crowdsourced worker to generate and annotate video scenes. For each person in each frame, workers had to answer whether the person had a false belief. Ultimately, 1213 scenarios were collected and passed quality control. This dataset can perform three tasks 1) who is mistaken 2) when are they mistaken to perform 3) joint reasoning on the problem of who and when.

BOSS\cite{duan2022boss} dataset is a benchmark for human belief prediction in object-context scenarios that provides 900 videos of social interactions, in addition to providing information on gaze, pose and contextual information in the environment. The two participants were required to accomplish a collaborative task by inferring and interpreting each other’s beliefs through non-verbal communication.

\subsection{Datasets for Intention Reasoning Tasks}
There exist more rich datasets in intention reasoning tasks, in forms of 2D grid, 3D animation and videos from human. 
In 2019, Carroll et al.\cite{strouse2021collaborating} published a dataset to evaluate whether AI can infer human intentions and cooperate with the human to complete a specific task in 2D grid, a popular game overcooked, in which players control chefs in a kitchen to cook and serve dishes. Carroll et al.\cite{strouse2021collaborating} created a web interface for the game to collect trajectories of humans. For each layout it gathered ~16 human-human trajectories (for a total of 18k environment time steps). It partitions the joint trajectories into two subsets, and split each trajectory into two single-agent trajectories. In recent years, many simulation platforms have been proposed, which can be used for human-robot interaction environment, some of which support 3D realistic scenes, with objects interacting and human-like agent. In Duan et al.\cite{duan2022surveyembodies}, the existing embodied AI platforms in recent years are compared and summarized. Compared with other platforms, there are fewer objects and it is a grid environment in overcooked, so the scene is relatively single. Only eight different scenarios are included in this dataset.

Puig et al.\cite{puig2020watch_WAH} presented a challenge called Watch And Help (WAH) for social perception and human-robot collaboration. Similar to overcooked tasks, AI needs to infer the intention of a human to accomplish the cooperation task. The task is a two-step process - watch stage and help stage. The AI first watches a video of Alice (a human like agent) successfully performing a household activity in a 3D scene, and then in a new scene, AI needs to help Alice to achieve the same goal with a minimum number of steps.
WAH designed a multi-agent simulation platform and created a benchmark containing 1011 tasks and 2 test sets. The dataset provided scene graph and behavior trajectories of each task, and we could use the emulator to generate videos of each task.

TIA \cite{wei2018and} is also a household dataset based on joint reasoning of tasks, intentions and attention. It provides 809 videos where a character performs a task and ask three questions: 1) where the human is looking - attention prediction; 2) why the human is looking there - intention prediction; and 3) what task the human is performing - task recognition. Each frame includes four types of data: the RGB image at resolution of 1920 $\times$ 1080, the depth image, the 3D human skeleton, and the egocentric RGB image at resolution of 1280 $\times$ 960 used for annotating the ground-truth attention points.

In contrast to overcooked, WAH and TIA datasets both provide large scale 3D data and involve more diverse scenarios in kitchen, living room and bath room. TIA dataset needs to answer the question of what is the intention of human. Overcooked pays attention to the interaction between AI and human, and it needs to infer human intention and cooperate with human, as a two-stage task.

\subsection{Datasets for Desire Reasoning Tasks}
At present, there are few datasets based on desire reasoning, and only one dataset focused on desire reasoning called MSED\cite{jia2022beyond_desire} proposed in 2022.

MSED is the first multi-modal dataset annotated with three sentiment classes, six emotion classes and six desire classes. The raw data was collected from online photo-sharing resources (i.e., Getty Image, Flickr and Twitter) by retrieving keywords with based on 16 basic desires theory, e.g., curiosity, romance, family, vengeance etc.
They crawled retrieved text-image posts on the first ten pages. After data filter, MSED dataset contains 9,190 text-image pairs, with 4683 samples for desire infer.
\subsection{Datasets for Joint Reasoning}
Baker et al.\cite{baker2017rational_BTOM} proposed an available dataset in 2D grid environment used for belief-desire joint reasoning. 73 scenarios are manually designed for the training of Bayesian network. Because this dataset is a small-scale designed to evaluate the reasoning ability of a Bayesian-based model and can not be used in deep-learing-based model.

BIB\cite{gandhi2021baby}and AGENT (Shu et al., 2021)\cite{shu2021agent} are large-scale and synthetic dataset of 3D video which are suitable for machine learning.
The scenes of BIB are conducted in 2D grid environment, and the dataset provides 2000 videos with both 2D and 3D versions of the stimuli rendered and scene configuration files describing the objects and agents present in the scene.
There are 8400 videos of 3D animations in AGENT of an agent moving under various physical constraints and interacting with various objects. There exists conceptual overlap with BIB while with distinct concepts such as unobserved constraints and cost-reward trade-off. In addition to that, AGNET is in 3D continuous space, so more diverse objects such as ramps, platforms, doors, and bridges, can be added to the scene while BIB contains scenes in maze-like environments that require more limited knowledge of physical constraints: mazes with walls. On the other hand, AGENT involves training on many different leave-out splits, where most splits have relatively minor differences between training and testing. AGENT and BIB provide complementary tools for benchmark machine agents’ core psychology reasoning, and relevant models could make use of both.

\section{Models}
The existing models on machine ToM are mainly divided into two directions: "model-based" and "cue-based". The "model-based" approaches build a causal relationship of  human mental states and then infer mental states from observations. The "cue-based" approaches assume that mentalizing is based on a direct mapping from behaviors to mental states. "Model-based" approaches are represented by Bayesian-based approaches and inverse reinforcement learning (IRL), while "cue-based" approaches are mainly represented by deep-learning-based approaches. In this article, we will introduce Bayesian-based approaches and deep-learning-based approaches in detail. Bayesian-based approaches have good explanatory because they model human mental states, but they can not fully reflect human mental states because they use a simplified ToM model. The deep-learning-based approaches are often applicable to the end-to-end training method. They directly map the action and mental states, and can get implicit mental states. In this section, we focus on two mainstream approaches and 
 the existing models are compared in Table \ref{model_compare} for the four approaches mentioned in this paper.
\begin{center}
\begin{table*}[h]
\normalsize

\renewcommand\arraystretch{1.3}
\caption{Models in theory of mind}
\centering
\resizebox{2\columnwidth}{!}{
\begin{tabular}{m{1cm}<{\raggedright}|m{2cm}<{\raggedright}|m{3cm}<{\raggedright}m{4cm}<{\raggedright}m{7cm}<{\raggedright}m{7cm}<{\raggedright}}
\hline
\textbf{Year} & \textbf{Model} & \textbf{Scene} & \textbf{Purpose} & \textbf{Advantage} & Limitation\\

\hline
\multicolumn{6}{c}{\textit{Deep-Learning-Based Approaches}}\\
\hline
2018 & ToMnet\cite{2018Machine}	& 2D grid& By learning from the past trajectory of an agent to infer mental states and predict the next actions in a new scene& Learning the mental state of agents with different behavior
	& A large amount of data is required for training\\
\hline	
2021	&ToMnet-G\cite{shu2021agent}	&3D scene&To predict the trajectory of an agent in a 3D scene &Extending ToMnet with GNN and pushes predictive scenarios from 2D to 3D
	&Lack of strong generalization ability within and across scenarios (once trained, only predicted under the same layout of the same scenario)	\\
\hline	
2020  & ToMnet+\cite{chuang2020using}	& 2D grid	& Inferring Human Social Interaction Preference by Using Computer Model to Observe the Behavior Data of the Third Person & Added preference inference based on ToMnet to infer human implicit preferences
&Predictions relying on learning from previous trajectory features and cannot be migrated \\
\hline
2022	&Trait-ToM\cite{nguyen2022learning}	&2D grid& Inferring other actors' mental states and goals by observing their past and current behavior.&	Proposing stable personality traits hold key prior information that affects mental states, achieved through 'quick weights', with significantly better performance than ToMnet	&Predictions rely on learning from previous trajectory features and cannot be migrated \\ 
\hline
\multicolumn{6}{c}{\textit{Bayesian-Based Approaches}}\\
\hline
2017&BTOM\cite{baker2017rational}	&Food-trucks 2D grid	&Joint reasoning of beliefs and desires  &	Extending the classical expectation-utility surrogate model to sequential actions in complex, partially observable domains	& Uniform distribution for belief hypothesis and unapplicable for general reasoning\\
\hline
2019&ATOM\cite{narang2019inferring}	& 3D multi-agent AR environment  & Inferring shared intention in a virtual environment& Explaining the underlying irrationality of human behavior and the dynamic nature of individual intentions & Inference ability and gaze response were limited to a single user, and there was no dependency between annotation and motion cues \\
\hline
2020&	MMDP\cite{2020Too}	& 2D grid & Inferring user intention and taking action to complete a cooperative task & With agents sharing tasks and conflict avoidance in multi-agent cooperation considered & Assumeing that humans are fully rational and does not take into account the false beliefs during cooperation\\
\hline

2019 & Speaker and listener\cite{2019A}	&Human-robot interaction experiment& Improved attention recognition method&Incorporating the intentional posture of its interacting partner into the intention infer&	Less performance gains and limited accuracy that can be extracted by modeling these advanced behaviors result in more narrow improvements\\
\hline
2020	&Thrive\cite{2020A}	&Human-robot interaction trust experiment&An integration of recent findings on potential mechanisms of trust in a computational model&	A demonstration of that ToM is important for building empathy trust, and uncovered trust-based learning mechanisms in children and robots.&Fewer influencing factors involved in trust-based learning\\
\hline
2022	&XAI\cite{yuan2022situ}	&Human-robot interaction game scene	&Proving the interpretability of mental theory&	Proof of the interpretability of mental theory by human-robot interaction	&Machines' overly rich interpretation of the decision-making process will be unacceptable to humans\\	
\hline
\multicolumn{6}{c}{\textit{Epistemic-Logic-Based Approach}}\\
\hline
2020	&CogTom\cite{grassiotto2021cogtom}&
Simple Text&  
First  order false belief detection
&False order beliefs can be detected.& Only symbolic language as input and first order belief detection\\
\hline
2020	&Lasse\cite{2020Implementing}&
images&  
First- and higher-order beliefs inference
&First- and second- order beliefs can be inferred dynamically.& Not incorporating other aspects of ToM such as intentions and desires.\\
\hline
\multicolumn{6}{c}{\textit{Inverse-Reinforcement-Learning-Based Approach}}\\
\hline
2018	&Dhruv\cite{2018An}&
CIRL game&  
Value alignment
&Reducing the complexity of IRL model and relax CIRL’s assumption of human rational-
ity.& A simple game setting with the low dimensional and discrete actions.\\
\hline

2021 &Tian\cite{Tian2021Learning}&
Pac-man game in 2D grid&  
Value alignment in multi-agent settings
&Considering intelligence levels when reasoning about ToM.& A discrete states and actions, and the high-cost computation complexity.\\
\hline

\end{tabular}
}
\label{model_compare}
\end{table*}

\end{center}
\subsection{Deep-Learning-Based Approaches}
\subsubsection{ToMnet}
\begin{figure}[htbp]
    \centering

    \subfloat[ToMnet architecture]{\includegraphics[width=0.60\textwidth,trim=0 0 350 220,clip]{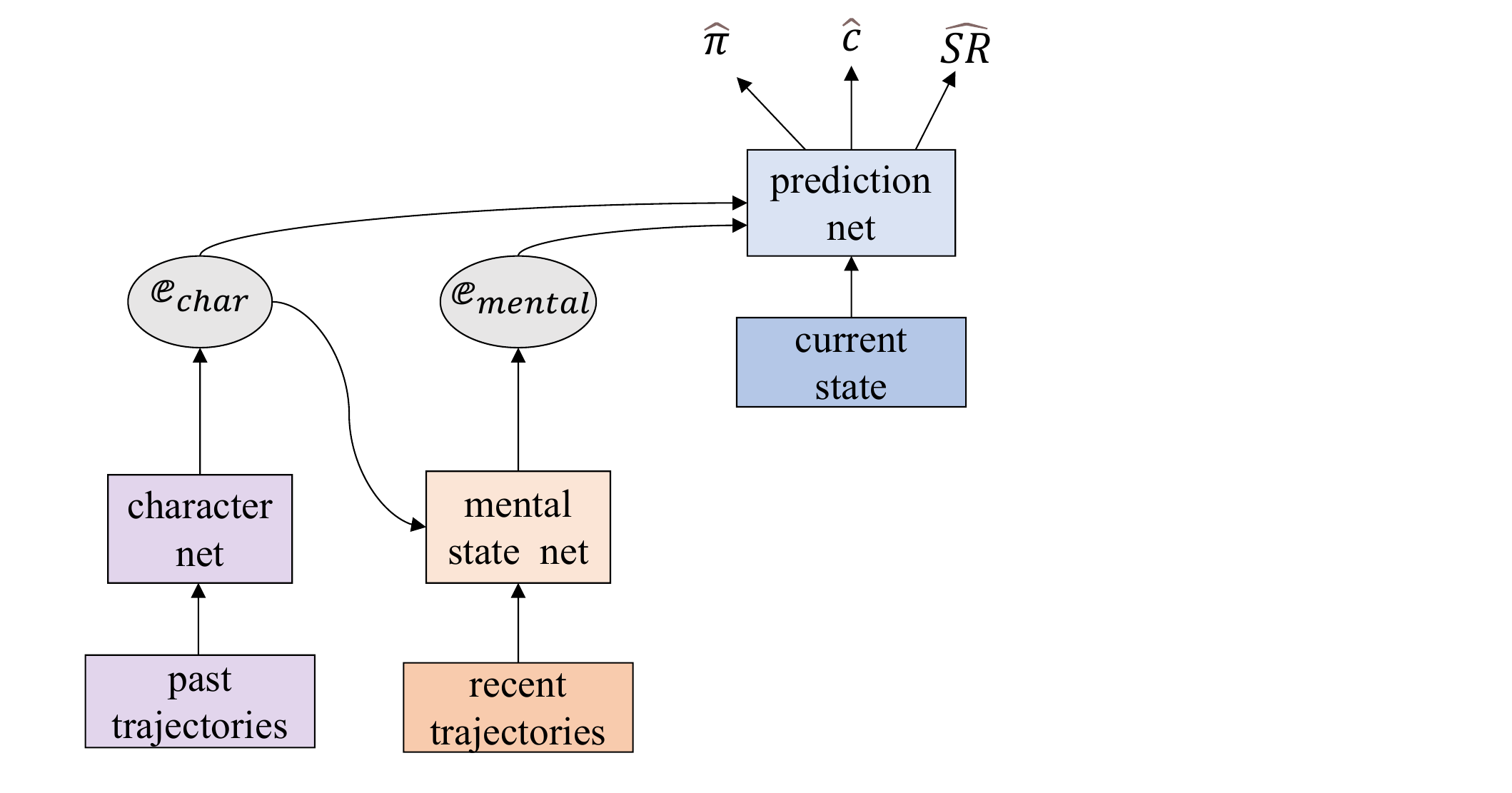}\label{fig: sub_figure1}%
    \label{fig_tom}}
    \hfil
    \subfloat[Trait-ToM architecture]{\includegraphics[width=0.49\textwidth,trim=40 30 50 40,clip]{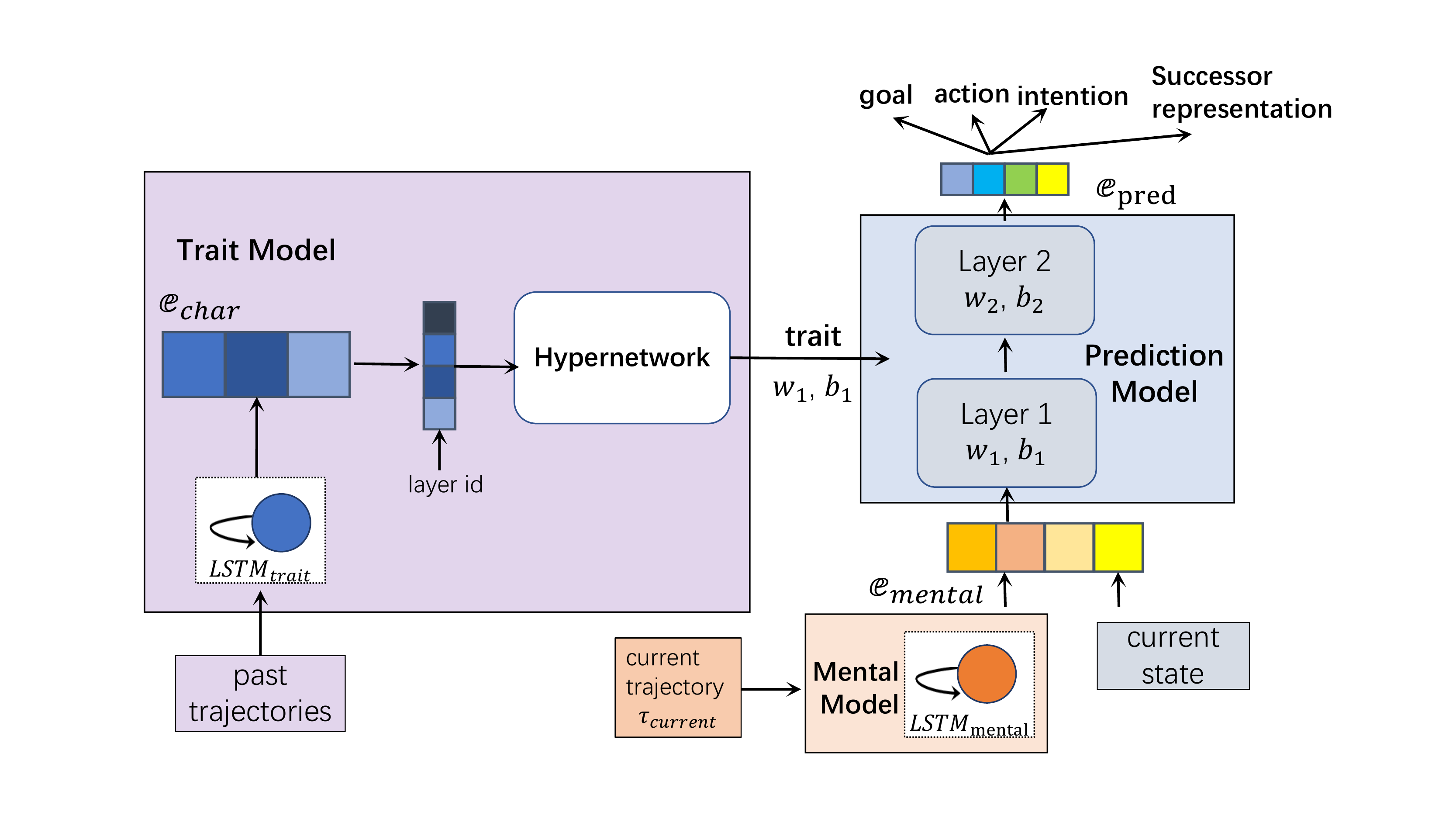}\label{fig: sub_figure3}
    \label{fig_trait}}
    \hfil
    \subfloat[ToMnet-G architecture]{\includegraphics[width=0.47\textwidth,trim=33 30 63 40,clip]{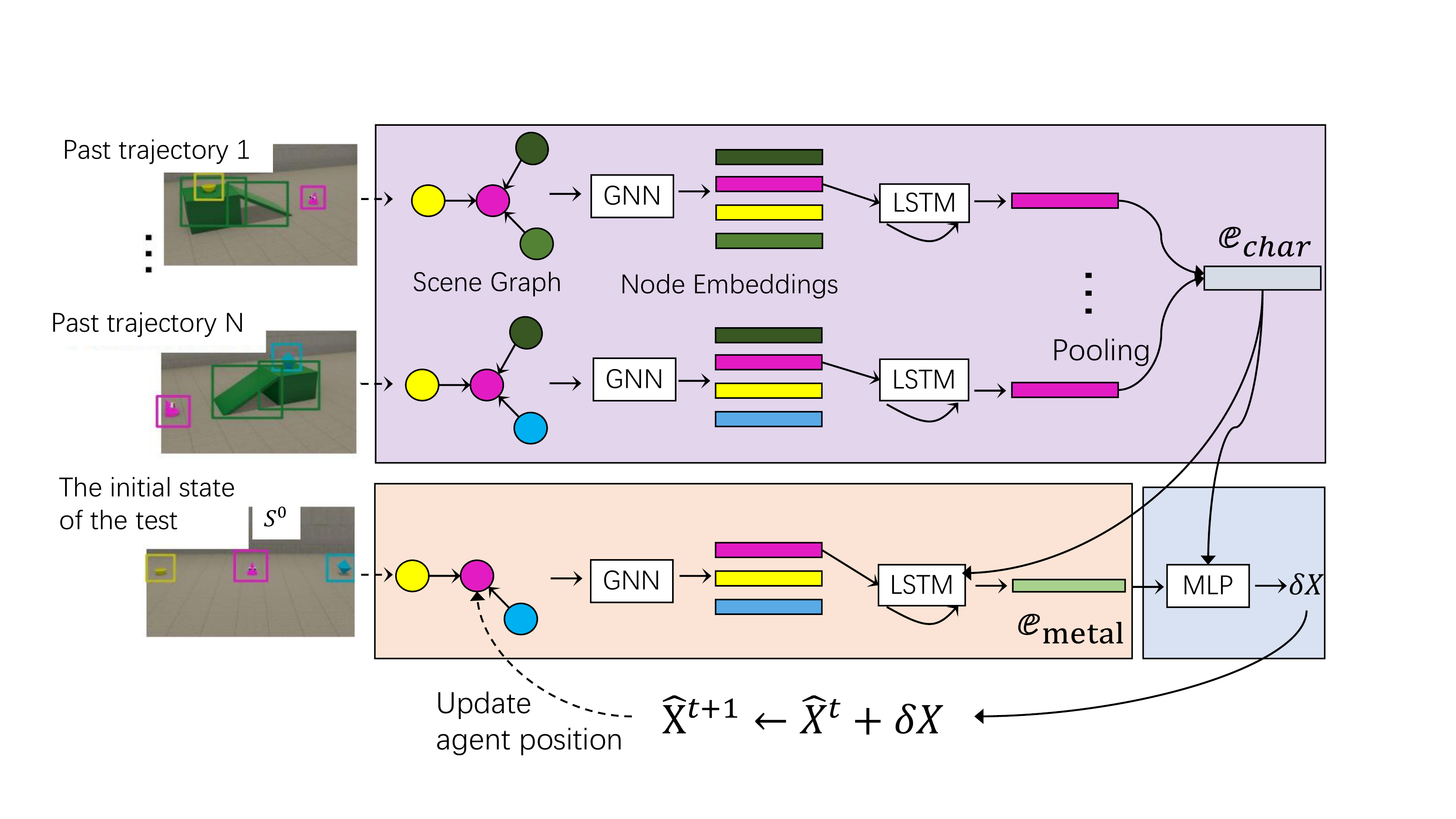}\label{fig: sub_figure4}
    \label{fig_tomnet_G}}
    \caption{The architecture of ToMnet \cite{2018Machine}, Trait-ToM\cite{nguyen2022learning} and ToMnet-G\cite{shu2021agent}. The three model are both consist of three modules: $character$ $net$ (the purple block), $mental$ $state$ $net$ (the orange block), and $prediction$ $net$ (the blue block). The $character$ $net$ parses an
    agent’s past trajectories. The $mental$ $state$ $net$ parses the agent’s trajectory on the current episode. These features from  $character$ $net$ and $mental$ $state$ $net$ are fed into the prediction net. The $prediction$ $net$ predictions about future behaviour in current state. ToMnet uses vector to represent agent's behaviour character and mental state while Trait-Tom uses fast weights.}
    \label{fig_model}
\end{figure}

DeepMind researchers have proposed a new neural network, ToMnet\cite{2018Machine} (ToMnet architecture
shown in Figure \ref{fig_tom}.), with the ability to understand its own mental state as well as that of the others. Agents with different behavioral characters collect colored squares of their targets and generate trajectories. ToMnet can observe the 
 history trajectories that represent ToM of an agent and predict the next action for the agent in a new scene. ToMnet is composed of three modules: \textit{a character net}, \textit{a mental state net}, and \textit{a prediction net}, each containing a number of small computational units and connections learned from experience, similar to the human brain. \textit{Character net} extracts the characters of agents based on their past actions represented by $e_{char}$ . \textit{Mental state net} understands the current mental state of agents $e_{mental}$; \textit{Prediction net} receives $e_{char}$ and $e_{mental}$, and speculates on the subsequent actions depending on the new situation and their characters. One of the limitations is that this model requires a large amount of data for training. Notwithstanding its limitations, it is shown that ToMnet can flexibly infer ToM over a range of different species of agents and has good generalization when mastering character over a new specie of agents that ToMnet has never seen.

 Chuang et al. proposed the ToMnet+ \cite{chuang2020using} adapted by ToMnet to infer the social network by observing the interactions between agents and targets. ToMnet+ tries to infer the current agent's social network from the past trajectories, and uses the inferred social network to help predict the best target for the agent in obtaining social support. Experiments have shown that using the ToMnet+ model to predict where an agent wants to go is over 80\% accurate.

\subsubsection{Trait-ToM}
Dung Nguyen et al. proposed the architecture of Trait-ToM \cite{nguyen2022learning} model. There are three sub-models (shown in Figure \ref{fig_trait}) where the character net is called \textit{trait model} and others are like ToMnet. Trait-ToM outputs intention as one of the label, while ToMnet does not. And unlike ToMnet which uses a vector to represent the character, Trait-ToM uses fast weights computed by \textit{trait model}. The fast-weight traits can directly modulate the function of the prediction net through multiplicative mechanisms and are higher in accuracy than ToMnet. In \textit{trait model}, a long-term and short-term memory network is used to maintain a dynamic state vector at each time step $t$ for every trajectory from state-action pair embedding. Finally, after averaging all past trajectories embedding, the hypernetwork is used to generate weights and biases, and feeds them to the prediction network as a parameter of this network with $e_{char}$. 
It is shown in Table \ref{tab:tom_and_trait_tom_result} that the Trait-ToM  outperforms the ToMnet in predicting the intention, especially in more realistic agent (S1 and S2).

\subsubsection{ToMnet-G}
Previous studies of deep-learning-based approaches of machine ToM  have not dealt with 3D scene. Tianmin Shu et al. extended ToMnet to predict the trajectory in a 3D scene, called ToMnet-G \cite{shu2021agent} (shown in Figture \ref{fig_tomnet_G}). ToMnet-G uses GNN to encode the states in a scene, using nodes to represent all objects in the scene and edges to represent the interaction between objects. Node information include objects types, bounding boxes, and colors. In order to better apply ToMnet-G to partially observable scene, it reconstructs the initial transitions of the past video and the test video. By sampling the trajectory, the network is trained using the mean square error loss of trajectory prediction. This model is suitable for 3D scene and predicts a continuous trajectory. The result shows that BIPaCk (a Bayesian-based model) outperforms ToMnet-G in almost all conditions, but BIPaCk requires an accurate reconstruction of the 3D state and a built-in model of the physical dynamics, which will not necessarily be available in real world scenes.

 \begin{table}[h]
 \caption{ToMnet and Trait-ToM Performance.} 
\label{tab:tom_and_trait_tom_result}
\tabcolsep=0.08cm
\renewcommand\arraystretch{1.2}

\scalebox{0.82}{
\begin{tabular}{llllllll}
\hline
\multirow{3}{*}{Stream} & \multirow{3}{*}{Model}     & \multicolumn{2}{l}{Random-Actor}       & \multicolumn{2}{l}{Hypo-Actor}         & \multicolumn{2}{l}{Hypo-Actor}            \\
                        &                            & \multicolumn{2}{l}{(Full Observation)} & \multicolumn{2}{l}{(Full Observation)} & \multicolumn{2}{l}{(Partial Observation)} \\ \cline{3-8} 
                        &                            & Action             & Intention         & Action             & Intention         & Action              & Intention           \\ \hline
\multirow{4}{*}{S1}     & \multirow{2}{*}{ToMnet}    & 37.8               & 53.21             & 54.26              & 53.43             & 49.65               & 51.94               \\
                        &                            & (0.64)             & (2.83)            & (0.94)             & (3.18)            & (0.96)              & (2.71)              \\ \cline{2-8} 
                        & \multirow{2}{*}{Trait-ToM} & 39.46              & 55.64             & 54.37              & 56.45             & 50.93               & 55.21               \\
                        &                            & (0.47)             & (3.02)            & (1.09)             & (2.48)            & (1.10)              & (2.43)              \\ \hline
\multirow{4}{*}{S2}     & \multirow{2}{*}{ToMnet}    & 32.48              & 36.6              & 50.29              & 46.59             & 44.3                & 43.96               \\
                        &                            & (1.71)             & (3.84)            & (2.32)             & (2.67)            & (2.45)              & (3.09)              \\ \cline{2-8} 
                        & \multirow{2}{*}{Trait-ToM} & 38.22              & 47.26             & 52.85              & 54.6              & 48.8                & 53.32               \\
                        &                            & (1.25)             & (2.77)            & (0.81)             & (2.35)            & (1.20)              & (2.30)              \\ \hline
\multirow{4}{*}{M}      & \multirow{2}{*}{ToMnet}    & 39.3               & 88.14             & 59.55              & 75.92             & 54.13               & 74.71               \\
                        &                            & (0.34)             & (1.05)            & (1.46)             & (2.35)            & (1.23)              & (1.83)              \\ \cline{2-8} 
                        & \multirow{2}{*}{Trait-ToM} & \textbf{40.84}     & \textbf{90.95}    & \textbf{60.53}     & \textbf{82.1}     & \textbf{55.52}      & \textbf{79.85}      \\
                        &                            & \textbf{(0.42)}    & \textbf{(0.57)}   & \textbf{(2.15)}    & \textbf{(1.52)}   & \textbf{(1.71)}     & \textbf{(1.11)}     \\ \hline
\multicolumn{8}{p{37em}}{\small $\bullet$ The two models are trained with various training setting (M, S1 and S1, for more details, see \cite{nguyen2022learning}, Table 1. S1 and S2 are more realistic scenarios, e.g. the observer can only see one type of agent at a time. In the mixed setting (M), agents in one batch are i.i.d sampled from 32 types of agent.}

\end{tabular}
}

\end{table}

\subsection{Bayesian-Based Approaches}
\subsubsection{BToM}
\label{sec:btom}
\begin{figure}[htbp]
\centering
 \includegraphics[width=0.35\textwidth,trim=550 160 0 0,clip]{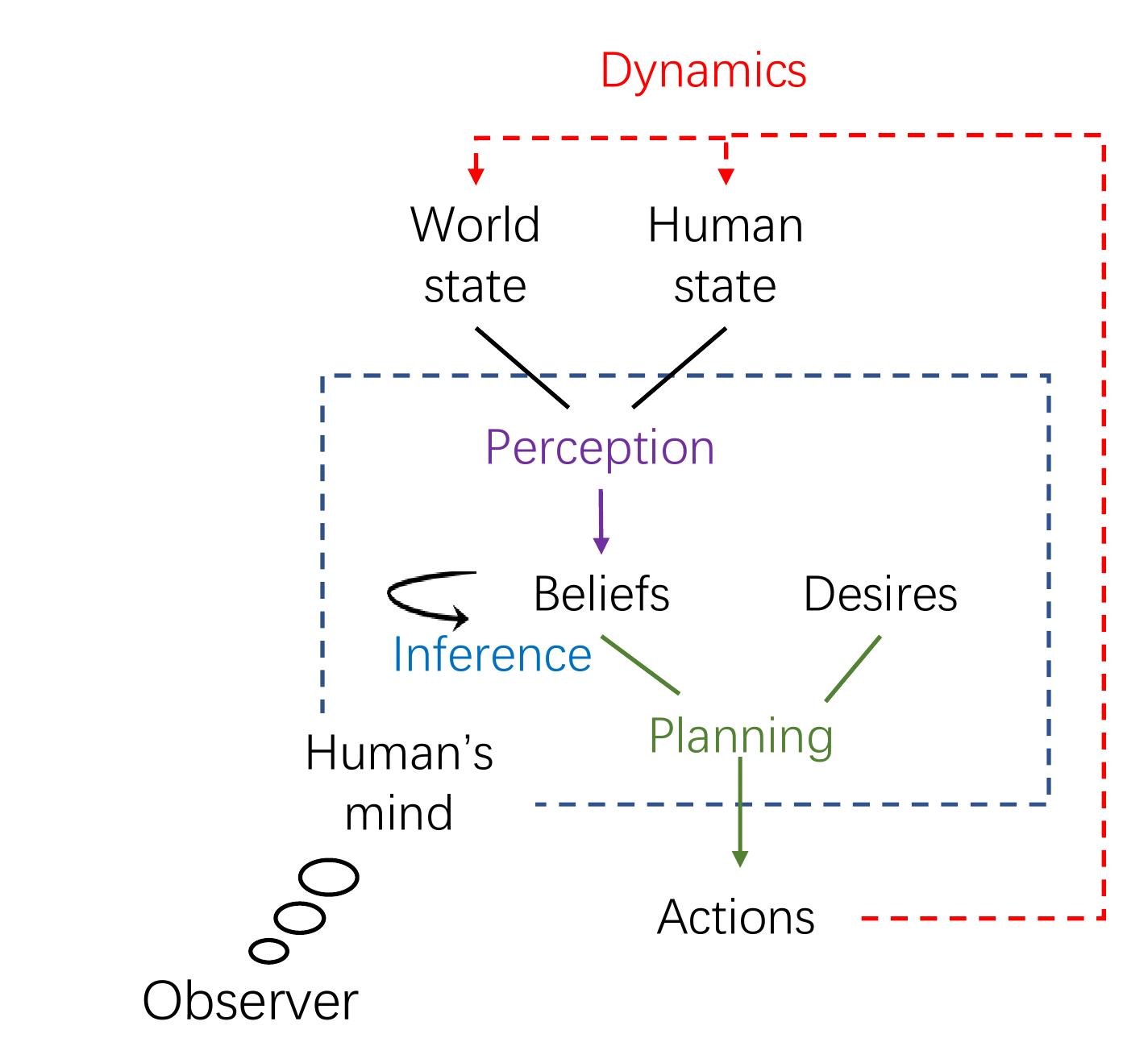}
\caption{A simplified ToM framework in BTOM \cite{baker2017rational}. Agent in BTOM, the observer views human as rational planner. Agent is observable to world state, human state and actions (outside the blue box), but unobservable to the mental states and process of inference ( inside the blue box).}
\label{fig:Folk-psychological schema for theory of mind}
\end{figure}

Baker et al.(2017) proposed the Bayesian Theory of Mind (BToM) \cite{baker2017rational_BTOM} model to explain human's beliefs and desires (More details about the experiment  can be found in Section \ref{sec:food_truck}).
BToM formalizes mentalization as Bayesian inference for a generative model of a rational agent. BToM uses POMDP to represent the agent’s planning and inference about the world. Figure \ref{fig:Folk-psychological schema for theory of mind} shows a simplified ToM framework of an agent-based for rational planning and state estimation, inspired by the classical theory of expected utility maximization decision making, but extended to the scene where agents plan uncertain actions in space and time due to incomplete information. 
POMDP captures three central causal principles of the core emphasized by the figure as follows. (I) A rational actor forms perceptions that are a rational function of the state of the world, their own state, and the nature of their perceptual apparatus - for a visual-guided actor, anything in their sight should be registered in their world model (Perception) . (II) The formation of beliefs, which are rational inferences based on the combination of their perceptions and their prior knowledge (Inference). (III) Plans for a reasonable sequence of actions - actions  based on their beliefs that can be expected to effectively and reliably achieve their aspirations (Planning).

BToM combines the POMDP generation model with the hypothesis space of candidate mental states and the prior space of these hypotheses. In the case of a given agent's behavior in a situational environment, Bayesian inference is performed on beliefs, desires, and perceptions.

\subsubsection{MMDP}
Sarah et al. \cite{2020Too} develop hierarchical planning architecture, a learning mechanism that enables agents to rapidly infer the intentions of other agents in a multi-agents cooperative task. Similar to BTOM, it views the agent decision-making process as a Markov decision process (MDP) and the model is a multi-agent MDP (MMDP). The architecture builds on two steps: high-level planning to infer others intention and decide its own intention, low-level planning to select action. In high-level planning, each agent views others as MDP and infers their intention by inverse planning. Low-level planning takes the sub-tasks selected by high-level planning and the next best action while modeling the movements of other agents. In this step, agents need to address two types of low-level coordination problems: (1) avoiding collisions while working on distinct sub-tasks, and (2) cooperating as necessary to solve a shared sub-task. When working on distinct sub-tasks, the agent treats others as static except for the agent who has collisions sub-task, so to simplify the model and task the best responds. When working a shared sub-task, it computes joint policies to view its and the other's decisions as a whole. While Bayesian delegation solves some aspects of commonsense coordination, one challenge is that when
agents share sub-task, they have no way of knowing when they have completed their individual "part" of the joint effort.

\subsubsection{AToM}
Sahil Narang et al. proposed a real-time algorithm called Agents with Theory of Mind (AToM) \cite{narang2019inferring} using Bayesian theory of mind, which can infer the hidden intentions of user avatars in a virtual environment based on observed proxemics and gaze-based cues. This method provides users with the first view in the multi-agent virtual environment. 
Each virtual agent independently perceives the user's social cues, using information such as theory of mind, gaze, actions, and expressions to infer the user's hidden intentions using Bayesian Theory of Mind based on the Markov assumption. Then the agent uses a velocity-space reasoning algorithm to compute velocity with respect
to visible entities in its neighborhood, including other agents and the
user and generates gaze-based response. 
The evaluation results show that the performance is significantly better than the previous methods. 
The previous methods are largely limited to the 2D scene with discrete actions. In contrast, this work focuses on role and multi-agent interactions in more complex environments with continuous action spaces. However, there are still some shortcomings in this model. For example, it defines intention simply based on observable features: the gaze and movement, which makes it easier to infer intentions.
\subsubsection{XAI}
\begin{figure}[htbp]
    \centering
    \includegraphics[width=0.54\textwidth,trim=48 120 0 0,clip]{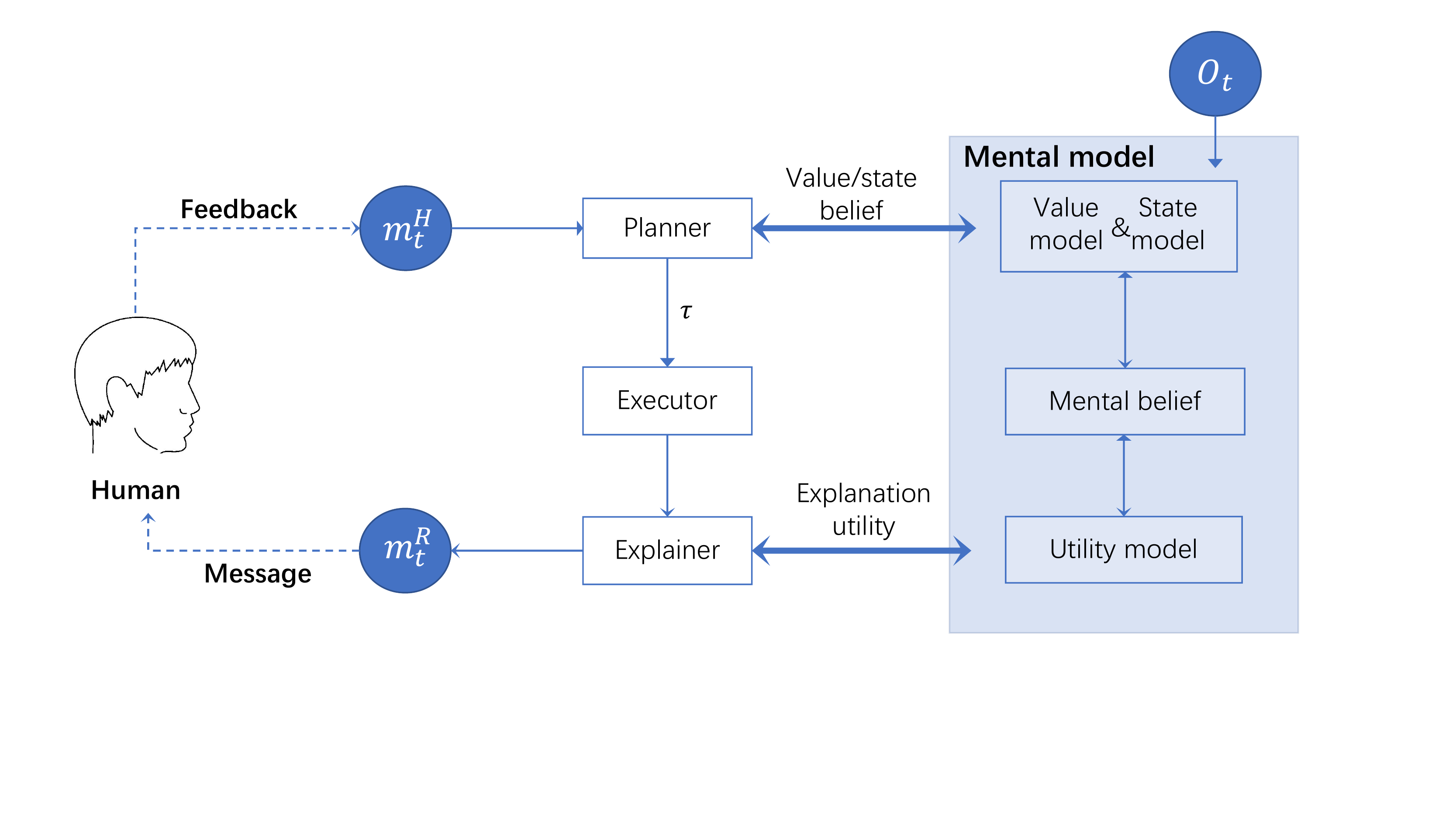}
    \caption{Algorithmic flow of the computational model. Given game observations and human feedback to previous proposals, the robots update their mental state and make new task plans. On the basis of the plans and current beliefs, new proposals and explanations are generated and sent to the human for feedback in the next round.}
    \label{fig:xai}
\end{figure}

Yuan, Luyao, et al. propose an explainable artificial intelligence (XAI) \cite{yuan2022situ} 
system. In the existing XAI system, human users’ active interactions or inputs to the system only influence explanations of robots’ decisions but rarely influence the model’s decision-making process.
In this study, the author designed a clever \textit{human-machine collaborative exploration} game to explore the process of value alignment between robots and humans (Figure \ref{fig:xai} shows the algorithmic flow of the computational model).
It is not difficult for the robots to simplify this game to POMDP setting and propose the optimal plan, if the robot scouts already know about the user’s value function. 
Thus, the difficulty is that robots need to align with the human values with sparsely supervised interactions.
Given that the user tends to provide helpful pedagogical feedback to facilitate alignment, the robots adopt a human-centric amelioration of iterative teacher-aware learning (ITAL)\cite{yuan2021iterative} to learn the value function.
It integrates two levels of ToM. Here, it is referred to the ability to infer humans' value from their actions as level-1 ToM and the comprehension of explicit information in users' feedback as level-2 ToM.
In each step of interaction, the robots present proposals and customized explanations, 
that reveal robots’ current estimation of human values and justify the proposed plan.
 After the users seed feedback for the proposal and explanations, the system formulates the overall generation as a hidden Markov model-based sequential generation process capable of adopting the temporal dynamics of the human user’s explanation utility.

\section{Summary}
This paper gives a comprehensive review on machine ToM in terms of beliefs, desires and intentions, covering the experiments, datasets and models in deep-learning-based approaches and Bayesian-based approaches. In this paper, the psychological experiments commonly used in recent years in machine ToM are described, and the variation of psychological experiments is summarized and compared. It is simple and easy to use grid experiment to reproduce psychological experiment. Thus, most of these experiments are based on grid experiments and psychological experiments as theoretical basis for model evaluation. Although many experiments are supported by psychological experiments, there is a lack of machine experimental paradigm for ToM. In addition, we find that the format and scale of datasets are constantly developing in recent years. The datasets extend from simple 2D gird to animation, and real 3D scenes and there are several large-scale datasets applicable for deep-learning-based approaches. There have not yet been a universal dataset that can be used for the evaluation of the three concepts. Finally, we compare the existing models. Both methods have their own advantages and limitations. At present, deep-learning-based approaches represented by the ToMnet framework are constantly proposed to solve different problems, but researchers still have to find a way to address the problems of the interpretability and the need for a large amount of training data for deep-learning-based model. Bayesian-based approaches also has some serious limitations. Once the networks are build, Bayesian-based approaches are difficult to extend and not as flexible as deep-learning-based approaches. Besides, there remains the issue of reflecting human's ToM in real life with a complex decision-making process instead of a simplified ToM.
Despite quite a few existing research, there is little work to compare the performance of the two approaches.

Generally, as a cognitive ability, ToM comprises many tasks in our daily life and can be formalized in different data formats, such as natural language, images, logical forms, etc. As a result, there are currently no clearly defined tasks and benchmark evaluation, leading to the difficulty that the proposed models can not be uniformly compared.

Solutions to this problem are now widely discussed. We argue that, on remedy is now to propose a standard task and then to build a large-scale dataset suitable for deep-learning-based approaches and covering ToM's multiple abilities assessment so to evaluate the models of different approaches in all aspects.

\clearpage





\ifCLASSOPTIONcompsoc
  \section*{Acknowledgments}
\else
  \section*{Acknowledgment}
\fi

This work is supported by National Key Research and Development Program of China (2021ZD0111000/2021ZD0111004), the Science and Technology Commission of Shanghai Municipality Grant (No. 21511100101, 22511105901). Xin Lin and Qin Ni are the corresponding authors.

\ifCLASSOPTIONcaptionsoff
  \newpage
\fi



\bibliographystyle{IEEEtran}
\bibliography{./main.bbl}

\begin{thebibliography}{10}
\providecommand{\url}[1]{#1}
\csname url@samestyle\endcsname
\providecommand{\newblock}{\relax}
\providecommand{\bibinfo}[2]{#2}
\providecommand{\BIBentrySTDinterwordspacing}{\spaceskip=0pt\relax}
\providecommand{\BIBentryALTinterwordstretchfactor}{4}
\providecommand{\BIBentryALTinterwordspacing}{\spaceskip=\fontdimen2\font plus
\BIBentryALTinterwordstretchfactor\fontdimen3\font minus
  \fontdimen4\font\relax}
\providecommand{\BIBforeignlanguage}[2]{{%
\expandafter\ifx\csname l@#1\endcsname\relax
\typeout{** WARNING: IEEEtran.bst: No hyphenation pattern has been}%
\typeout{** loaded for the language `#1'. Using the pattern for}%
\typeout{** the default language instead.}%
\else
\language=\csname l@#1\endcsname
\fi
#2}}
\providecommand{\BIBdecl}{\relax}
\BIBdecl

\bibitem{gandhi2021baby}
K.~Gandhi, G.~Stojnic, B.~M. Lake, and M.~R. Dillon, ``Baby intuitions
  benchmark (bib): Discerning the goals, preferences, and actions of others,''
  \emph{Advances in Neural Information Processing Systems}, vol.~34, pp.
  9963--9976, 2021.

\bibitem{shu2021agent}
T.~Shu, A.~Bhandwaldar, C.~Gan, K.~Smith, S.~Liu, D.~Gutfreund, E.~Spelke,
  J.~Tenenbaum, and T.~Ullman, ``Agent: A benchmark for core psychological
  reasoning,'' in \emph{International Conference on Machine Learning}.\hskip
  1em plus 0.5em minus 0.4em\relax PMLR, 2021, pp. 9614--9625.

\bibitem{duan2022surveyembodies}
J.~Duan, S.~Yu, H.~L. Tan, H.~Zhu, and C.~Tan, ``A survey of embodied ai: From
  simulators to research tasks,'' \emph{IEEE Transactions on Emerging Topics in
  Computational Intelligence}, 2022.

\bibitem{weston2015towards_facebook_bAbi}
J.~Weston, A.~Bordes, S.~Chopra, A.~M. Rush, B.~Van~Merri{\"e}nboer, A.~Joulin,
  and T.~Mikolov, ``Towards ai-complete question answering: A set of
  prerequisite toy tasks,'' \emph{arXiv preprint arXiv:1502.05698}, 2015.

\bibitem{grant2017can}
E.~Grant, A.~Nematzadeh, and T.~L. Griffiths, ``How can memory-augmented neural
  networks pass a false-belief task?''

\bibitem{nematzadeh2018evaluating}
A.~Nematzadeh, K.~Burns, E.~Grant, A.~Gopnik, and T.~L. Griffiths, ``Evaluating
  theory of mind in question answering,'' \emph{arXiv preprint
  arXiv:1808.09352}, 2018.

\bibitem{le2019revisiting}
M.~Le, Y.-L. Boureau, and M.~Nickel, ``Revisiting the evaluation of theory of
  mind through question answering,'' in \emph{Proceedings of the 2019
  Conference on Empirical Methods in Natural Language Processing and the 9th
  International Joint Conference on Natural Language Processing
  (EMNLP-IJCNLP)}, 2019, pp. 5872--5877.

\bibitem{eysenbach2016mistaken}
B.~Eysenbach, C.~Vondrick, and A.~Torralba, ``Who is mistaken?'' \emph{arXiv
  preprint arXiv:1612.01175}, 2016.

\bibitem{duan2022boss}
J.~Duan, S.~Yu, N.~Tan, L.~Yi, and C.~Tan, ``Boss: A benchmark for human belief
  prediction in object-context scenarios,'' \emph{arXiv preprint
  arXiv:2206.10665}, 2022.

\bibitem{wei2018and}
P.~Wei, Y.~Liu, T.~Shu, N.~Zheng, and S.-C. Zhu, ``Where and why are they
  looking? jointly inferring human attention and intentions in complex tasks,''
  in \emph{Proceedings of the IEEE conference on computer vision and pattern
  recognition}, 2018, pp. 6801--6809.

\bibitem{strouse2021collaborating}
D.~Strouse, K.~McKee, M.~Botvinick, E.~Hughes, and R.~Everett, ``Collaborating
  with humans without human data,'' \emph{Advances in Neural Information
  Processing Systems}, vol.~34, pp. 14\,502--14\,515, 2021.

\bibitem{puig2020watch_WAH}
X.~Puig, T.~Shu, S.~Li, Z.~Wang, Y.-H. Liao, J.~B. Tenenbaum, S.~Fidler, and
  A.~Torralba, ``Watch-and-help: A challenge for social perception and human-ai
  collaboration,'' \emph{arXiv preprint arXiv:2010.09890}, 2020.

\bibitem{jia2022beyond_desire}
A.~Jia, Y.~He, Y.~Zhang, S.~Uprety, D.~Song, and C.~Lioma, ``Beyond emotion: A
  multi-modal dataset for human desire understanding,'' in \emph{Proceedings of
  the 2022 Conference of the North American Chapter of the Association for
  Computational Linguistics: Human Language Technologies}, 2022, pp.
  1512--1522.

\bibitem{baker2017rational_BTOM}
C.~L. Baker, J.~Jara-Ettinger, R.~Saxe, and J.~B. Tenenbaum, ``Rational
  quantitative attribution of beliefs, desires and percepts in human
  mentalizing,'' \emph{Nature Human Behaviour}, vol.~1, no.~4, pp. 1--10, 2017.

\bibitem{1985Does}
S.~Baron-Cohen, A.~M. Leslie, and U.~Frith, ``Does the autistic child have a
  "theory of mind" ?'' \emph{Cognition}, vol.~21, no.~1, pp. 0--46, 1985.

\bibitem{2018Evaluating}
A.~Nematzadeh, K.~Burns, E.~Grant, A.~Gopnik, and T.~Griffiths, ``Evaluating
  theory of mind in question answering,'' 2018.

\bibitem{2018Machine}
N.~C. Rabinowitz, F.~Perbet, H.~F. Song, C.~Zhang, S.~Eslami, and M.~Botvinick,
  ``Machine theory of mind,'' 2018.

\bibitem{2020Online}
Z.~X. Tan, J.~L. Mann, T.~Silver, J.~B. Tenenbaum, and V.~K. Mansinghka,
  ``Online bayesian goal inference for boundedly-rational planning agents,''
  2020.

\bibitem{overcook}
\BIBentryALTinterwordspacing
D.~Strouse, K.~R. McKee, M.~M. Botvinick, E.~Hughes, and R.~Everett,
  ``Collaborating with humans without human data,'' in \emph{Advances in Neural
  Information Processing Systems 34: Annual Conference on Neural Information
  Processing Systems 2021, NeurIPS 2021, December 6-14, 2021, virtual},
  M.~Ranzato, A.~Beygelzimer, Y.~N. Dauphin, P.~Liang, and J.~W. Vaughan, Eds.,
  2021, pp. 14\,502--14\,515. [Online]. Available:
  \url{https://proceedings.neurips.cc/paper/2021/hash/797134c3e42371bb4979a462eb2f042a-Abstract.html}
\BIBentrySTDinterwordspacing

\bibitem{2019A}
J.~Lee, S.~Fei, and C.~L. Breazeal, ``A bayesian theory of mind approach to
  nonverbal communication,'' in \emph{2019 14th ACM/IEEE International
  Conference on Human-Robot Interaction (HRI)}, 2019.

\bibitem{2020A}
M.~Patacchiola and A.~Cangelosi, ``A developmental cognitive architecture for
  trust and theory of mind in humanoid robots,'' \emph{IEEE Transactions on
  Cybernetics}, vol.~PP, no.~99, pp. 1--13, 2020.

\bibitem{1998Infants}
A.~Woodward, ``Infants selectively encode the goal object of an actor's
  reach,'' \emph{Cognition}, vol.~69, no.~1, p.~1, 1998.

\bibitem{2003One}
G.~Csibra, S.~Biro, O.~Koos, and G.~Gergely, ``One-year-old infants use
  teleological representations of actions productively,'' \emph{Cognitive
  Science}, vol.~27, no.~1, pp. 111--133, 2003.

\bibitem{2017Ten}
S.~Liu, T.~D. Ullman, J.~B. Tenenbaum, and E.~S. Spelke, ``Ten-month-old
  infants infer the value of goals from the costs of actions,'' \emph{Science},
  vol. 358, no. 6366, pp. 1038--1041, 2017.

\bibitem{1999Infants}
A.~L. Woodward, ``Infants’ ability to distinguish between purposeful and
  non-purposeful behaviors,'' \emph{Infant Behavior and Development}, 1999.

\bibitem{BURESH2007287}
\BIBentryALTinterwordspacing
J.~S. Buresh and A.~L. Woodward, ``Infants track action goals within and across
  agents,'' \emph{Cognition}, vol. 104, no.~2, pp. 287--314, 2007. [Online].
  Available:
  \url{https://www.sciencedirect.com/science/article/pii/S001002770600151X}
\BIBentrySTDinterwordspacing

\bibitem{Ren1992The}
Renée, Baillargeon, Amy, Needham, Julie, and Devos, ``The development of young
  infants' intuitions about support,'' \emph{Early Development \& Parenting},
  1992.

\bibitem{doi:10.1177/0956797612457395}
\BIBentryALTinterwordspacing
R.~M. Scott and R.~Baillargeon, ``Do infants really expect agents to act
  efficiently? a critical test of the rationality principle,''
  \emph{Psychological Science}, vol.~24, no.~4, pp. 466--474, 2013, pMID:
  23470355. [Online]. Available: \url{https://doi.org/10.1177/0956797612457395}
\BIBentrySTDinterwordspacing

\bibitem{2005Pulling}
J.~A. Sommerville and A.~L. Woodward, ``Pulling out the intentional structure
  of action: the relation between action processing and action production in
  infancy,'' \emph{Cognition}, vol.~95, no.~1, pp. 1--30, 2005.

\bibitem{Gy1995Taking}
G.~Gergely, Z.~Nádasdy, G.~Csibra, and S.~Bíró, ``Taking the intentional
  stance at 12 months of age,'' \emph{Cognition}, vol.~56, no.~2, pp. 165--193,
  1995.

\bibitem{premack1978does}
D.~Premack and G.~Woodruff, ``Does the chimpanzee have a theory of mind?''
  \emph{Behavioral and brain sciences}, vol.~1, no.~4, pp. 515--526, 1978.

\bibitem{tager2000componential}
H.~Tager-Flusberg and K.~Sullivan, ``A componential view of theory of mind:
  evidence from williams syndrome,'' \emph{Cognition}, vol.~76, no.~1, pp.
  59--90, 2000.

\bibitem{leslie1987pretense}
A.~M. Leslie, ``Pretense and representation: The origins of" theory of
  mind.",'' \emph{Psychological review}, vol.~94, no.~4, p. 412, 1987.

\bibitem{wellman2001meta}
H.~M. Wellman, D.~Cross, and J.~Watson, ``Meta-analysis of theory-of-mind
  development: The truth about false belief,'' \emph{Child development},
  vol.~72, no.~3, pp. 655--684, 2001.

\bibitem{baillargeon1987young}
R.~Baillargeon, ``Young infants' reasoning about the physical and spatial
  properties of a hidden object,'' \emph{Cognitive Development}, vol.~2, no.~3,
  pp. 179--200, 1987.

\bibitem{spelke1990principles}
E.~S. Spelke, ``Principles of object perception,'' \emph{Cognitive science},
  vol.~14, no.~1, pp. 29--56, 1990.

\bibitem{kim1992infants}
I.~K. Kim and E.~S. Spelke, ``Infants' sensitivity to effects of gravity on
  visible object motion.'' \emph{Journal of Experimental Psychology: Human
  Perception and Performance}, vol.~18, no.~2, p. 385, 1992.

\bibitem{oakes1990infant}
L.~M. Oakes and L.~B. Cohen, ``Infant perception of a causal event,''
  \emph{Cognitive Development}, vol.~5, no.~2, pp. 193--207, 1990.

\bibitem{needham1993intuitions}
A.~Needham and R.~Baillargeon, ``Intuitions about support in 4.5-month-old
  infants,'' \emph{Cognition}, vol.~47, no.~2, pp. 121--148, 1993.

\bibitem{xu1996infants}
F.~Xu and S.~Carey, ``Infants’ metaphysics: The case of numerical identity,''
  \emph{Cognitive psychology}, vol.~30, no.~2, pp. 111--153, 1996.

\bibitem{liszkowski200612}
U.~Liszkowski, M.~Carpenter, T.~Striano, and M.~Tomasello, ``12-and
  18-month-olds point to provide information for others,'' \emph{Journal of
  cognition and development}, vol.~7, no.~2, pp. 173--187, 2006.

\bibitem{saxe2004understanding}
R.~Saxe, S.~Carey, and N.~Kanwisher, ``Understanding other minds,'' \emph{Annu.
  Rev. Psychol}, vol.~55, pp. 87--124, 2004.

\bibitem{wimmer1983beliefs}
H.~Wimmer and J.~Perner, ``Beliefs about beliefs: Representation and
  constraining function of wrong beliefs in young children's understanding of
  deception,'' \emph{Cognition}, vol.~13, no.~1, pp. 103--128, 1983.

\bibitem{hala1997all}
S.~Hala and J.~Carpendale, ``All in the mind: Children’s understanding of
  mental life,'' \emph{The development of social cognition}, vol.~2, pp.
  189--239, 1997.

\bibitem{callaghan2005synchrony}
T.~Callaghan, P.~Rochat, A.~Lillard, M.~L. Claux, H.~Odden, S.~Itakura,
  S.~Tapanya, and S.~Singh, ``Synchrony in the onset of mental-state reasoning:
  Evidence from five cultures,'' \emph{Psychological Science}, vol.~16, no.~5,
  pp. 378--384, 2005.

\bibitem{colonnesi2005emergence}
C.~Colonnesi, ``The emergence of a theory of mind from infancy to childhood,''
  \emph{Tesi di Dottorato}, 2005.

\bibitem{wellman2000developing}
H.~M. Wellman and K.~H. Lagattuta, ``Developing understandings of mind.'' 2000.

\bibitem{steele2003brief}
S.~Steele, R.~M. Joseph, and H.~Tager-Flusberg, ``Brief report: Developmental
  change in theory of mind abilities in children with autism,'' \emph{Journal
  of autism and developmental disorders}, vol.~33, no.~4, pp. 461--467, 2003.

\bibitem{onishi200515}
K.~H. Onishi and R.~Baillargeon, ``Do 15-month-old infants understand false
  beliefs?'' \emph{science}, vol. 308, no. 5719, pp. 255--258, 2005.

\bibitem{sperberattribution}
D.~Sperber and S.~Caldi, ``Attribution of beliefs by 13-month-old infants.''

\bibitem{surian2007attribution}
L.~Surian, S.~Caldi, and D.~Sperber, ``Attribution of beliefs by 13-month-old
  infants,'' \emph{Psychological science}, vol.~18, no.~7, pp. 580--586, 2007.

\bibitem{knudsen201218}
B.~Knudsen and U.~Liszkowski, ``18-month-olds predict specific action mistakes
  through attribution of false belief, not ignorance, and intervene
  accordingly,'' \emph{Infancy}, vol.~17, no.~6, pp. 672--691, 2012.

\bibitem{trauble2010early}
B.~Tr{\"a}uble, V.~Marinovi{\'c}, and S.~Pauen, ``Early theory of mind
  competencies: Do infants understand others’ beliefs?'' \emph{Infancy},
  vol.~15, no.~4, pp. 434--444, 2010.

\bibitem{southgate2010seventeen}
V.~Southgate, C.~Chevallier, and G.~Csibra, ``Seventeen-month-olds appeal to
  false beliefs to interpret others’ referential communication,''
  \emph{Developmental science}, vol.~13, no.~6, pp. 907--912, 2010.

\bibitem{surian2012will}
L.~Surian and A.~Geraci, ``Where will the triangle look for it? attributing
  false beliefs to a geometric shape at 17 months,'' \emph{British Journal of
  Developmental Psychology}, vol.~30, no.~1, pp. 30--44, 2012.

\bibitem{clements1994implicit}
W.~A. Clements and J.~Perner, ``Implicit understanding of belief,''
  \emph{Cognitive development}, vol.~9, no.~4, pp. 377--395, 1994.

\bibitem{buttelmann2009eighteen}
D.~Buttelmann, M.~Carpenter, and M.~Tomasello, ``Eighteen-month-old infants
  show false belief understanding in an active helping paradigm,''
  \emph{Cognition}, vol. 112, no.~2, pp. 337--342, 2009.

\bibitem{knudsen2012eighteen}
B.~Knudsen and U.~Liszkowski, ``Eighteen-and 24-month-old infants correct
  others in anticipation of action mistakes,'' \emph{Developmental science},
  vol.~15, no.~1, pp. 113--122, 2012.

\bibitem{georgeff1998belief}
M.~Georgeff, B.~Pell, M.~Pollack, M.~Tambe, and M.~Wooldridge, ``The
  belief-desire-intention model of agency,'' in \emph{International workshop on
  agent theories, architectures, and languages}.\hskip 1em plus 0.5em minus
  0.4em\relax Springer, 1998, pp. 1--10.

\bibitem{hadfield2016cooperative}
D.~Hadfield-Menell, S.~J. Russell, P.~Abbeel, and A.~Dragan, ``Cooperative
  inverse reinforcement learning,'' \emph{Advances in neural information
  processing systems}, vol.~29, 2016.

\bibitem{jara2019theory}
J.~Jara-Ettinger, ``Theory of mind as inverse reinforcement learning,''
  \emph{Current Opinion in Behavioral Sciences}, vol.~29, pp. 105--110, 2019.

\bibitem{woodward1998infants}
A.~L. Woodward, ``Infants selectively encode the goal object of an actor's
  reach,'' \emph{Cognition}, vol.~69, no.~1, pp. 1--34, 1998.

\bibitem{1998Planning}
L.~P.~K. A, M.~L.~L. B, and A.~R.~C. C, ``Planning and acting in partially
  observable stochastic domains,'' \emph{Artificial Intelligence}, vol. 101,
  no. 1–2, pp. 99--134, 1998.

\bibitem{2017What}
B.~F. Malle and S.~T. Magar, ``What kind of mind do i want in my robot?:
  Developing a measure of desired mental capacities in social robots,'' in
  \emph{the Companion of the 2017 ACM/IEEE International Conference}, 2017.

\bibitem{Slaughter2015Theory}
Slaughter and Virginia, ``Theory of mind in infants and young children: A
  review,'' \emph{Australian Psychologist}, vol.~50, no.~3, p. 169–172, 2015.

\bibitem{Carruthers2013Mindreading}
Carruthers and Peter, ``Mindreading in infancy,'' \emph{Mind \& Language},
  vol.~28, no.~2, pp. 141--172, 2013.

\bibitem{2017Replications}
L.~J. Powell, K.~Hobbs, A.~Bardis, S.~Carey, and R.~Saxe, ``Replications of
  implicit theory of mind tasks with varying representational demands,''
  \emph{Cognitive Development}, p. S0885201417300461, 2017.

\bibitem{2018How}
S.~D?Rrenberg, H.~Rakoczy, and U.~Liszkowski, ``How (not) to measure infant
  theory of mind: Testing the replicability and validity of four non-verbal
  measures,'' \emph{Cognitive Development}, p. S0885201417300448, 2018.

\bibitem{Kulke2018Is}
Kulke, Louisa, von, Duhn, Britta, Schneider, Dana, Rakoczy, and Hannes, ``Is
  implicit theory of mind a real and robust phenomenon? results from a
  systematic replication study,'' \emph{Psychological science: a journal of the
  American Psychological Society}, vol.~29, no.~6, pp. 888--900, 2018.

\bibitem{10.2307/1130707}
\BIBentryALTinterwordspacing
A.~Gopnik and V.~Slaughter, ``Young children's understanding of changes in
  their mental states,'' \emph{Child Development}, vol.~62, no.~1, pp. 98--110,
  1991. [Online]. Available: \url{http://www.jstor.org/stable/1130707}
\BIBentrySTDinterwordspacing

\bibitem{langley2022theory}
C.~Langley, B.~I. Cirstea, F.~Cuzzolin, and B.~J. Sahakian, ``Theory of mind
  and preference learning at the interface of cognitive science, neuroscience,
  and ai: A review,'' \emph{Frontiers in Artificial Intelligence}, p.~62, 2022.

\bibitem{2004Reasoning}
Y.~Dimopoulos, A.~C. Kakas, and L.~Michael, ``Reasoning about actions and
  change in answer set programming,'' \emph{lecture notes in computer science},
  2004.

\bibitem{2020Implementing}
L.~Dissing and T.~Bolander, ``Implementing theory of mind on a robot using
  dynamic epistemic logic,'' in \emph{International Joint Conference on
  Artificial Intelligence}, 2020.

\bibitem{2018The}
Y.~Chen, A.~Saffidine, and C.~Schwering, ``The complexity of limited belief
  reasoning—the quantifier-free case,'' 2018.

\bibitem{nguyen2020cognitive}
T.~N. Nguyen and C.~Gonzalez, ``Cognitive machine theory of mind,'' Carnegie
  Mellon University, Tech. Rep., 2020.

\bibitem{yuan2022situ}
L.~Yuan, X.~Gao, Z.~Zheng, M.~Edmonds, Y.~N. Wu, F.~Rossano, H.~Lu, Y.~Zhu, and
  S.-C. Zhu, ``In situ bidirectional human-robot value alignment,''
  \emph{Science robotics}, vol.~7, no.~68, p. eabm4183, 2022.

\bibitem{baker2017rational}
C.~L. Baker, J.~Jara-Ettinger, R.~Saxe, and J.~B. Tenenbaum, ``Rational
  quantitative attribution of beliefs, desires and percepts in human
  mentalizing,'' \emph{Nature Human Behaviour}, vol.~1, no.~4, pp. 1--10, 2017.

\bibitem{chuang2020using}
Y.-S. Chuang, H.-Y. Hung, E.~Gamborino, J.~O.~S. Goh, T.-R. Huang, Y.-L. Chang,
  S.-L. Yeh, and L.-C. Fu, ``Using machine theory of mind to learn agent social
  network structures from observed interactive behaviors with targets,'' in
  \emph{2020 29th IEEE International Conference on Robot and Human Interactive
  Communication (RO-MAN)}.\hskip 1em plus 0.5em minus 0.4em\relax IEEE, 2020,
  pp. 1013--1019.

\bibitem{nguyen2022learning}
D.~Nguyen, P.~Nguyen, H.~Le, K.~Do, S.~Venkatesh, and T.~Tran, ``Learning
  theory of mind via dynamic traits attribution,'' \emph{arXiv preprint
  arXiv:2204.09047}, 2022.

\bibitem{narang2019inferring}
S.~Narang, A.~Best, and D.~Manocha, ``Inferring user intent using bayesian
  theory of mind in shared avatar-agent virtual environments,'' \emph{IEEE
  transactions on visualization and computer graphics}, vol.~25, no.~5, pp.
  2113--2122, 2019.

\bibitem{grassiotto2021cogtom}
F.~Grassiotto and P.~D.~P. Costa, ``Cogtom: A cognitive architecture
  implementation of the theory of mind.'' in \emph{ICAART (2)}, 2021, pp.
  546--553.

\bibitem{baron1997mindblindness}
S.~Baron-Cohen, \emph{Mindblindness: An essay on autism and theory of
  mind}.\hskip 1em plus 0.5em minus 0.4em\relax MIT press, 1997.

\bibitem{yuan2021iterative}
L.~Yuan, D.~Zhou, J.~Shen, J.~Gao, J.~L. Chen, Q.~Gu, Y.~N. Wu, and S.-C. Zhu,
  ``Iterative teacher-aware learning,'' \emph{Advances in Neural Information
  Processing Systems}, vol.~34, pp. 29\,231--29\,245, 2021.

\bibitem{2020Too}
R.~E. Wang, S.~A. Wu, J.~A. Evans, J.~B. Tenenbaum, D.~C. Parkes, and
  M.~Kleiman-Weiner, ``Too many cooks: Coordinating multi-agent collaboration
  through inverse planning,'' 2020.

\bibitem{2017People}
A.~Jern, C.~G. Lucas, and C.~Kemp, ``People learn other people's preferences
  through inverse decision-making,'' \emph{Cognition}, vol. 168, pp. 46--64,
  2017.

\bibitem{2018An}
D.~Malik, ``An efficient, generalized bellman update for cooperative inverse
  reinforcement learning,'' 2018.

\bibitem{Tian2021Learning}
R.~Tian, M.~Tomizuka, and L.~Sun, ``Learning human rewards by inferring their
  latent intelligence levels in multi-agent games: A theory-of-mind approach
  with application to driving data,'' 2021.

\end{thebibliography}

%







\begin{IEEEbiography}[{\includegraphics[width=1in,height=1.25in,clip,keepaspectratio]{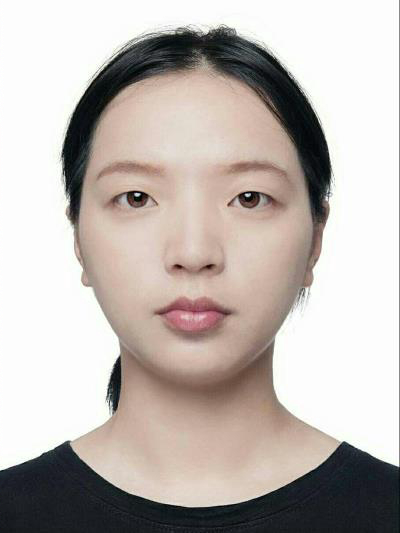}}]{Mao Yuanyuan}
 is currently pursuing for a master’s degree in computer science and technology with the College of  , East China Normal University. Her research interests include machine learning and machine cognitive assessment.Contact her at 51215901051@stu.ecnu.edu.cn.
\end{IEEEbiography}

\begin{IEEEbiography}[{\includegraphics[width=1in,height=1.25in,clip,keepaspectratio]{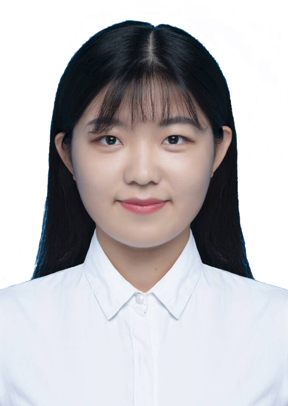}}]{Shuang Liu}
 is currently pursuing for a master’s degree in computer technology with the College of Information, Mechanical and Electrical Engineering, Shanghai Normal University. Her research interests include machine learning and machine cognitive assessment.Contact her at 1000513395@smail.shnu.edu.cn.
\end{IEEEbiography}
\begin{IEEEbiography}[{\includegraphics[width=1in,height=1.25in,clip,keepaspectratio]{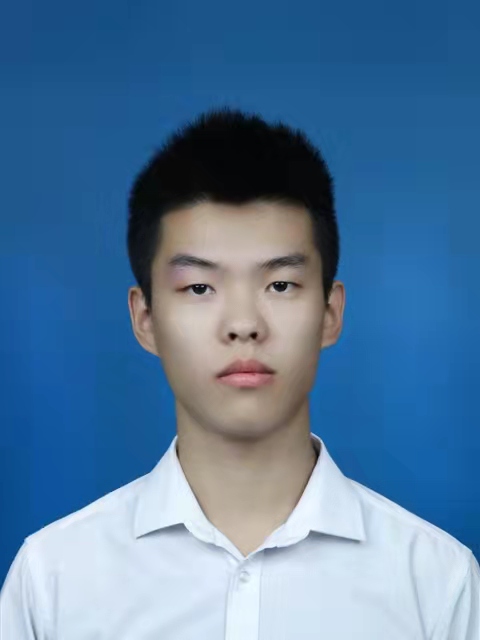}}]{Pengshuai Zhao}
 is currently pursuing for a master’s degree in computer technology with the College of Information, Mechanical and Electrical Engineering, Shanghai Normal University. His research interests include machine learning and machine cognitive assessment.Contact he at spz45683968@gmail.com

\end{IEEEbiography}

\begin{IEEEbiography}[{\includegraphics[width=1in,height=1.25in,clip,keepaspectratio]{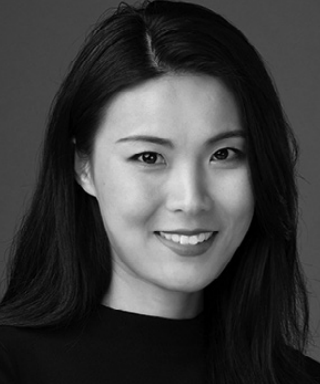}}]{Qin Ni}
is an associate professor in college of information, mechanical and electrical engineering of Shanghai Normal University, PR China. She received the PhD degree in system and information service from the Technical University of Madrid, Spain in 2016 with the scholarship from China Scholarship Council. She took the post of visiting scholar at the University of Ulster, U.K. in 2015. Her research interests include educational data mining, adaptive learning and smart environment. Contact her at niqin@shnu.edu.cn.

\end{IEEEbiography}

\begin{IEEEbiography}[{\includegraphics[width=1in,height=1.25in,clip,keepaspectratio]{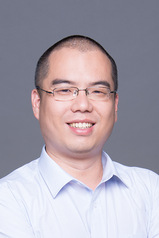}}]{Xin Lin}
received the BEng and PhD degrees both in computer science and engineering from Zhejiang University, China. He is currently an professor in the Department of Computer Science, East China Normal University. His research interests include knowledge base, machine TOM and graph computing.
\end{IEEEbiography}

\begin{IEEEbiography}[{\includegraphics[width=1in,height=1.25in,clip,keepaspectratio]{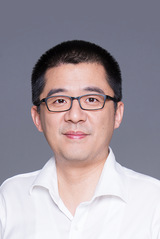}}]{Liang He}
received his bachelor’s degree and PhD degree from the Department of Computer Science and Technology, East China Normal University, Shanghai, China. He is now a professor and the associate dean of the School of Computer Science and Technology, East China Normal University. His current research interest includes knowledge processing, user behavior analysis, and context-aware computing. He has been awarded the Star of the Talent in Shanghai. He is also a council member of the Shanghai Computer Society,
a member of the Academic Committee, the director of the technical committee of Shanghai Engineering Research Center of Intelligent Service Robot, and a technology foresight expert of the Shanghai Science and Technology in focus areas.
\end{IEEEbiography}

\end{document}